\documentclass[twoside]{article}

\usepackage[accepted]{aistats2026}
\usepackage[round]{natbib}

\usepackage{graphicx} % Required for inserting images
\usepackage{amsmath}
\usepackage{amsthm}
\usepackage{amssymb}
\usepackage{xcolor}
\usepackage{hyperref}
\usepackage{listings}
\usepackage{nicematrix}
\usepackage{booktabs}
\usepackage{enumitem}
\usepackage[nameinlink,capitalize,noabbrev]{cleveref}
\usepackage{multirow}

\definecolor{codegreen}{rgb}{0,0.6,0}
\definecolor{codegray}{rgb}{0.5,0.5,0.5}
\definecolor{codepurple}{rgb}{0.58,0,0.82}
\definecolor{backcolour}{rgb}{0.96,0.96,0.96}

\lstdefinestyle{mystyle}{
    backgroundcolor=\color{backcolour},   
    commentstyle=\color{codegreen},
    keywordstyle=\color{magenta},
    numberstyle=\tiny\color{codegray},
    stringstyle=\color{codepurple},
    basicstyle=\ttfamily\footnotesize,
    breakatwhitespace=false,
    breaklines=true,
    captionpos=b,
    keepspaces=true,                 
    numbers=left,                    
    numbersep=5pt,                  
    showspaces=false,                
    showstringspaces=false,
    showtabs=false,                  
    tabsize=2
}

\newtheorem*{remark}{Remark}

\begin{document}

\twocolumn[

\aistatstitle{Structured Matrix Scaling for Multi-Class Calibration}

\aistatsauthor{ Eugène Berta \And David Holzmüller \And Michael I. Jordan \And  Francis Bach}

\aistatsaddress{Inria - ENS,\\PSL Research University \And Inria \And Inria - ENS,\\PSL Research University \And Inria - ENS,\\PSL Research University}]

\begin{abstract}
Post-hoc recalibration methods are widely used to ensure that classifiers provide faithful probability estimates.
We argue that parametric recalibration functions based on logistic regression can be motivated from a simple theoretical setting for both binary and multi-class classification.
This insight motivates the use of more expressive calibration methods beyond standard temperature scaling.
For multi-class calibration however, a key challenge lies in the increasing number of parameters introduced by more complex models, often coupled with limited calibration data, which can lead to overfitting.
Through extensive experiments, we demonstrate that the resulting bias-variance tradeoff can be effectively managed by structured regularization, robust preprocessing and efficient optimization.
The resulting methods lead to substantial gains over existing logistic-based calibration techniques.
We provide efficient and easy-to-use open-source implementations of our methods, making them an attractive alternative to common temperature, vector, and matrix scaling implementations.
\end{abstract}

\section{INTRODUCTION}

In multi-output classification, we aim to build a classifier~$f$ that predicts a categorical outcome $Y \in \{ 1, \dots, k \}$ from a feature vector $X \in \mathcal{X}$.
We assume that $(X,Y)$ are drawn independently from an underlying distribution $\mathcal{D}$, and we are given a dataset of feature vectors and labels $(x_i, y_i)_{1 \leq i \leq n}$ sampled from~$\mathcal{D}$.

Many modern classifiers make continuous predictions in the probability simplex $f(X) = p \in \Delta_k$, with $\Delta_k = \{ p \in [0,1]^k \,|\, \sum_{i=1}^k p_i = 1 \}$. Each component $p_i$ represents the model's confidence that the true class is $Y=i$ and can be interpreted as the estimated probability of that event occurring.
This probabilistic interpretation is valid only if the predicted probabilities align with the true underlying probabilities, a property known as \emph{calibration}.
Formally, a model~$f$ is calibrated if $f(X) = \mathbb{P}(Y\,|\,f(X))$ almost surely for $(X,Y) \sim \mathcal{D}$. When this holds, the prediction made by~$f$ for a new observation can be understood as a categorical distribution over the possible outcomes, which makes forecasts readily interpretable.

Unfortunately, classifiers trained on finite data, even with proper losses like cross-entropy, often exhibit significant miscalibration \citep{zadrozny2002transforming, guo2017calibration}.
A common remedy is \emph{post-hoc calibration}, where a recalibration function $g$ is applied to the output of the initial classifier $f$ to better align its predictions with true probabilities.
This is typically done by specifying a parametric model $g_\theta$ and finding parameters $\theta$ that minimize a loss function on a reserved calibration set:
$$
\min_\theta \frac{1}{n_\text{cal}} \sum_{i=1}^{n_\text{cal}} \ell(g_\theta \circ f(x_i), y_i) \, .
$$
Critically, calibration data is often scarce, with ${n_\text{cal} \ll n}$, as we generally prefer to use most of the available data to train the classifier~$f$.
Thus, a fundamental tradeoff arises that is analogous to bias-variance tradeoffs in regression: an overly simple $g_\theta$ may be insufficient to correct the miscalibration, while with an overly complex model one risks overfitting the calibration set, potentially degrading both calibration and predictive performance.
The required complexity of $g_\theta$ depends on both the amount of calibration data available and the nature of the miscalibration itself.

\paragraph{A gap between theory and practice.}
To understand the fundamental nature of this tradeoff, we begin by analyzing an idealized setting. Consider in particular a binary classification problem with Gaussian class-conditional data; that is, $\mathbb{P}(X \mid Y =0)$ and $\mathbb{P}(X \mid Y=1)$ are both normally distributed with different means and variances. Consider a logistic regression classifier, $f(X) = \sigma(w^\top X)$, where $\sigma(z) = (1 + e^{-z})^{-1}$ denotes the sigmoid function. Surprisingly, even in this simple scenario the optimal recalibration function is \emph{quadratic} in the logit:
\[
\mathbb{P}(Y=1 \mid f(X)=s) = \sigma\big( a \sigma^{-1}(s)^2 + b \sigma^{-1}(s) + c \big) \, ,
\]
where $\sigma^{-1}(s) = \log \frac{s}{1-s}$ is the logit function (see \cref{appendix:BinaryCase} for the full derivation).
In contrast, widely used methods such as temperature scaling \citep{guo2017calibration}
\[
g_b(s) = \sigma(b \sigma^{-1}(s)) \, ,
\]
or Platt scaling \citep{platt1999probabilistic} on the logits (also known as beta[a=b] calibration \citep{kull2017beta})
\[
g_{b,c}(s) = \sigma(b \sigma^{-1}(s) + c) \, ,
\]
are linear or affine in the logits. Thus, even in this idealized setting, existing approaches are fundamentally inadequate for recalibrating the classifier $f$.
In \cref{appendix:BinaryCase} we demonstrate through extensive experiments that post-hoc calibration quadratic in the logits outperforms the existing linear and affine methods, showing that this theoretical gap translates in practice.

This complexity gap becomes even more pronounced in the multi-class setting, on which we focus in the following.
Our analysis shows that logistic regression on multi-class Gaussian data requires post-hoc calibration with a quadratic softmax model (\cref{sec:multiclassGaussianModel})—far more complex than the commonly used temperature or vector scaling \citep{guo2017calibration}.
Previous attempts at using more expressive models such as matrix scaling have been hampered by overfitting \citep{guo2017calibration}, leading practitioners to employ simpler but potentially inadequate methods.

\paragraph{Our approach.}
In this paper, we show that adequate regularization enables expressive calibration functions while avoiding overfitting. We propose structured regularization schemes that adapt the complexity of post-hoc calibration to the amount of calibration data available, defaulting to simpler forms when data is scarce, and exploiting more expressive structures when sufficient data is available (\cref{sec:StructuredRegularization}).
This approach allows the model to capture a wide range of miscalibration patterns while remaining robust to overfitting.

We select default regularization parameters to provide strong out-of-the-box performance (\cref{appendix:HyperparameterSearch}). Our methods also allow for fine-tuning for specific applications.
Through extensive experiments, we demonstrate that our methods yield substantial improvement in post-hoc calibration (\cref{sec:MulticlassExperiments}).

\paragraph{Contributions.}  
Our work introduces three main contributions:
\begin{enumerate}[topsep=0pt, itemsep=0pt]
    \item We provide theoretical motivation showing that even simple classification problems require calibration functions of higher complexity than commonly assumed.
    \item We introduce structured regularization schemes that balance expressiveness and overfitting, enabling safe use of more powerful logistic calibration models. Thanks to carefully chosen hyperparameters, optimizers, and preprocessing, our methods are robust to various forms of miscalibration, alleviating the need for hyperparameter tuning.
    \item We release efficient implementations in the open-source package \texttt{probmetrics}\footnote{\url{https://github.com/dholzmueller/probmetrics}}, which provides faster and more accurate calibration out of the box than existing logistic methods, making our methods practical replacements for commonly used techniques.
\end{enumerate}

\section{MOTIVATING CALIBRATION WITH LOGISTIC REGRESSION} \label{sec:multiclassGaussianModel}

In the following, we denote by $S:\mathbb{R}^k \rightarrow \Delta_k$ the softmax function $S(x)_i = \frac{\exp x_i}{\sum_{j=1}^k \exp x_j}$, by $I_k$ the $k \times k$ identity matrix, and by $\mathbf{1}_k$ the $k$-dimensional constant vector equal to $1$ on every dimension.

We will see that we can motivate the use of multinomial logistic regression (also known as softmax regression) for post-hoc calibration from a simple setting.

Let us assume that $\mathcal{X} = \mathbb{R}^d$ and that the class-conditional distributions are multivariate normal distributions: $\forall i \in \{ 1, \cdots, k \},  \mathbb{P}(X \mid Y=i) = \mathcal{N}(\mu_i, \Sigma_i)$ with mean vectors $(\mu_i)_{1 \leq i \leq k} \in \mathbb{R}^d$ and covariance matrices $(\Sigma_i)_{1 \leq i \leq k} \in \mathbb{S}_+^{d \times d}$. Define the softmax regression classifier, $f(X) = S(WX)$, where $W \in \mathbb{R}^{k \times d}$ is a weight matrix mapping the feature vector $X \in \mathbb{R}^d$ to a logit vector $WX \in \mathbb{R}^k$. The distribution of logits $WX$ in this case is normally distributed for each class:
$$
\mathbb{P}(WX \mid Y=i) = \mathcal{N}(W \mu_i, W \Sigma_i W^\top) \, .
$$
In the multi-class case, a difficulty comes from the fact that the softmax function is not an injection. Indeed, for any scalar $c$, $S(z + c \mathbf{1}_k) = S(z)$. For this reason, the events $WX = z$ and $S(WX) = S(z)$ are not equivalent and (unlike in the binary case) the conditional distribution $\mathbb{P}(Y \mid WX)$ does not immediately gives us $\mathbb{P}(Y \mid f(X))$.

To overcome this difficulty, we use the $k \times k$ centering matrix $C_k = I_k - \frac{1}{k} \mathbf{1}_k \mathbf{1}_k^\top$. The nullspace of $C_k$ is exactly the invariant space of the Softmax: $\{ c \mathbf{1}_k \,|\, c \in \mathbb{R} \}$.
Consequently, for any centered vector $z \in \mathbb{R}^k$ with $C_k z = z$, the events $C_k W X = z$ and $S(WX) = S(z)$ are equivalent. For a given class $i$, the distribution of centered logits, $C_k WX$, is also a Gaussian: $\mathbb{P}(C_k WX \mid Y=i) = \mathcal{N}(C_kW\mu_i, C_kW\Sigma_iW^\top C_k^\top)$. However, $C_k$ is not full rank, so the induced covariance matrix $C_k W \Sigma_i W^\top C_k^\top$ is singular. This means that the distribution is degenerate and has no density with respect to the $k$-dimensional Lebesgue measure. Fortunately, the disintegration theorem yields a well-defined density with respect to the Lebesgue measure when restricted to the $(k-1)$-dimensional subspace of centered vectors.
Denoting by $\Sigma^+$ the generalized inverse, $\,|\,\Sigma\,|\,_+$ the pseudo determinant, $m_i := C_k W \mu_i$ the class means, $\sigma_i := C_k W \Sigma_i W^\top C_k^\top$ the class covariances and $\pi_i$ the class probabilities for all $i \in \{ 0, \cdots, k \} $, the class-conditional density writes
\begin{align*}
    \mathbb{P}(C_kWX \! = \! x \mid Y \!= \! i) = \frac{ \! \exp \! \big( \! - \! \frac{1}{2} (x \! - \! m_i)^\top \! \sigma_i^+ (x \! - \! m_i) \big)}{\sqrt{\,|\,2 \pi \sigma_i\,|\,_+}} .
\end{align*}

Applying Bayes' theorem (see \cref{appendix:Proofs}), we obtain  that the conditional distribution of $Y$ is
$$
    \mathbb{P}(Y \mid C_k WX = x) =  S(x^\top \mathbf{A} x + B x + C) \, ,
$$
where $\mathbf{A} \in \mathbb{R}^{k \times k \times k}$ is a three-dimensional array defined by $\mathbf{A}[i,:,:] = - \frac{1}{2} \sigma_i^+$ so that $(x^\top \mathbf{A} x)_i = - \frac{1}{2} x^T \sigma_i^+ x$, where $B$ is a matrix in $\mathbb{R}^{k \times k}$ defined by $B_i = m_i^\top \sigma_i^+$ and where $C$ is a vector in $\mathbb{R}^k$ defined by $C_i = - \frac{1}{2} m_i^\top \sigma_i^+ m_i + \log(\pi_i \,|\,\sigma_i\,|\,_+^{-1/2})$.

These calculations result in the following expression as a model for calibration:
$$
\mathbb{P}(Y | f(X) \! = \! s) = S(S^{-1}(s)^\top \mathbf{A} S^{-1}(s) + B S^{-1}(s) + C) \, ,
$$
where $S^{-1}$ denotes the one-to-one mapping from probabilities to centered logit vectors:
$S^{-1}(x) = C_k [ \log(x_1), \cdots, \log(x_k) ]^\top$.
Considering the post-hoc calibration function
$$
g(x) = S(S^{-1}(x) \mathbf{A} S^{-1}(x) + B S^{-1}(x) + C) \, ,
$$the same proof as the binary case (in \cref{appendix:BinaryCase}) allows us to conclude that $g \circ f$ is calibrated,
$$
\mathbb{P}(Y \mid g \circ f(X) = p) = p \, .
$$
Notice that if all class-conditional covariance matrices $(\Sigma_i)_{1 \leq i \leq k}$ are equal, the quadratic term in the softmax is constant and can thus be ignored. In this case, matrix scaling suffices for recalibration.

\begin{remark}
    In our example, the softmax function is well specified for calibration because the class-conditional logit distributions are Gaussian.
    However, performing post-hoc calibration with softmax regression does not correspond to assuming that the class-conditional distributions are normally distributed.
    There exists a wide range of distributions for which the conditional is logistic, as discussed in \cite{jordan1995logistic} and \cite{boken2021appropriateness}.
    We refer the reader to \citet{bishop2006pattern} for details on when the softmax model is well specified.
\end{remark}

\section{SOFTMAX CALIBRATION FUNCTIONS} \label{sec:StructuredRegularization}

The analysis in \cref{sec:multiclassGaussianModel} motivates the use of a logistic model as a post-hoc calibration function.
In practice however, we cannot compute coefficients $\mathbf{A}$, $B$ and $C$ directly.
Real-world data are not normally distributed and the classifier need not be logistic regression, so logistic calibration is not necessarily well specified.
Still, hopefully there exist coefficients of the softmax model~$g$ for which $g \circ f$ is approximately calibrated, or at least better calibrated than $f$.
To find such coefficients, we resort to empirical risk minimization.
Using our reserved calibration set, $(x_i, y_i)_{1 \leq i \leq n_\text{cal}}$, we solve the following optimization problem:
\begin{equation} \label{eq:multiclassLogisticRegression}
    \min_g \frac{1}{n_\text{cal}} \sum_{i=1}^{n_\text{cal}} \ell(g \circ f(x_i), y_i) \, ,
\end{equation}
where $g$ is parametrized by the coefficients of the logistic model. $\ell$ is typically the logloss, which makes the problem convex and thus allows us to find the optimal coefficients easily.

In the literature, different choices based on the softmax model have been considered for $g$:
\begin{itemize}
    \item Linear scaling, or temperature scaling (TS) \citep{guo2017calibration}: $g(x) = S( \alpha S^{-1}(x))$, also applies in the multi-class case, $g$ is parametrized only by a scalar coefficient $\alpha \in \mathbb{R}$.
    \item Vector scaling (VS) \citep{guo2017calibration}: $g(x) = S( \text{diag}(v) S^{-1}(x) + b)$, $g$ is parametrized by an intercept vector $b \in \mathbb{R}^k$ and a $k \times k$ diagonal weight matrix with diagonal $v \in \mathbb{R}^k$.
    \item Matrix scaling (MS) \citep{guo2017calibration}: $g(x) = S( M S^{-1}(x) + b)$, $g$ is parametrized by an intercept vector $b \in \mathbb{R}^k$ and a full weight matrix $M \in \mathbb{R}^{k \times k}$.
    \item To the best of our knowledge, the quadratic softmax model, $g(x) = S(S^{-1}(x)^\top \mathbf{Q} S^{-1}(x) + M S^{-1}(x) + b)$, with $\mathbf{Q}$ a $k \times k \times k$ array of parameters, has not been studied in the literature. We will argue that this is an oversight.
\end{itemize}

\begin{remark}
    In \cref{sec:multiclassGaussianModel} we introduced $S^{-1}$ as the one-to-one mapping between probabilities and centered logits $S^{-1}(x) = C_k [\log(x_1), \cdots, \log(x_k)]^\top$. In practice previous work does not use centering and the multi-class logit function is simply $S^{-1}(x) = [\log(x_1), \cdots, \log(x_k)]^\top$.
\end{remark}

We cannot minimize the population risk $\mathbb{E}_\mathcal{D}[\ell(g \circ f(X), Y)]$ directly, but only an empirical estimate on the calibration set \eqref{eq:multiclassLogisticRegression}. In the multi-class case, our proposed calibration function $g$ can have a large number of parameters ($k^3 + k^2 +k$ for the full quadratic model).
The risk of overfitting to the calibration set, which can result in poor calibration but also degraded performance, thus needs to be considered carefully. 

To navigate this risk, we take the point of view that post-hoc calibration is akin to solving a second probabilistic classification problem, with $k$ classes and $k$ features (the log probabilities), and under the constraint that the size of the calibration set $n_\text{cal}$ is generally small. Well-established machine learning tools can help us strike the right balance.
We note that \cite{kull2019dirichlet} provide a first example of this approach, by introducing ad hoc regularization on the intercept and off-diagonal coefficients of matrix scaling, yielding encouraging empirical results. Our approach introduces regularization from first principles, and we will demonstrate that this theoretically grounded approach can result in significant improvements.

It is already known from empirical studies that matrix scaling without regularization can lead to drastic overfitting on the calibration set \citep{guo2017calibration, kull2019dirichlet}. For this reason, and to keep the number of parameters reasonable, we decide not to explore a quadratic model, which leaves us with three models of increasing complexity: linear scaling (one parameter), vector scaling ($2k$ parameters) and matrix scaling ($k(k+1)$ parameters). Depending on the number of classes and the amount of data, we would like our model to adaptively choose between these different layers of complexity. In particular, we wish to build a single model that can replicate the behavior of the three models and never perform worse than any single model.

One common issue when using regularization for post hoc calibration is that the impact of regularization heavily depends on the scale of the logits, and thus on the confidence of the initial model.
To circumvent this, we propose applying temperature scaling first, to bring the logits to a common scale.
On top of it, we fit the following \emph{structured matrix scaling} (SMS) calibration function
$$
 g_\text{SMS}(x) \! = \! S \big( ( I_k + \mathrm{diag}(v) + (\mathbf{1}_k \mathbf{1}_k^\top \! - \! I_k) \odot M) S^{-1}(x) + b \big) ,
$$
where $\odot$ denotes the element-wise or Hadamard product.
Notice that we apply the off-diagonal mask $\mathbf{1}_k \mathbf{1}_k^\top - I_k$ coefficient-wise on the full matrix $M$ so that its diagonal coefficients are inactive and there is no overlap with the effect of the vector parameter $v$.
Graphically, our function matches traditional softmax regression on the logits, using a weight matrix $W$ with the following structure:
\begin{equation*}
\NiceMatrixOptions{xdots/shorten=0.5em}
W = \begin{pNiceMatrix}
1 + v_1 & M_{1,2} & \Ldots &M_{1,k}\\
M_{2,1} & 1 + v_2 & \Ddots & \Vdots \\
\Vdots & \Ddots & \Ddots & M_{k-1, k} \\
M_{k, 1} & \Ldots & M_{k, k-1} & 1 + v_k
\end{pNiceMatrix} \, .
\end{equation*}

This hierarchical parameter structure gives us the freedom to apply different regularization strengths on different parameter groups.
%The linear parameter $\alpha$ is a global scaling parameter that applies to all classes.
The diagonal vector $v$ allows each class to have a separate temperature parameter.
The off-diagonal parameters, $(\mathbf{1}_k \mathbf{1}_k^\top - I_k) \odot M$, allow for more complex inter-class dependencies.
Finally the intercept vector~$b$ allows for class-specific intercepts.
To learn the parameters of $g_\text{SMS}$, we solve the convex optimization problem
\begin{multline}\label{eq:FittingSMS}
    \min_{b, v, M} \frac{1}{n_\text{cal}} \sum_{i=1}^{n_\text{cal}} \ell(g_\text{SMS}(x_i), y_i) + \lambda_b \frac{k^\rho}{n_\text{cal}^\tau}\|b\|_\delta\\
    + \lambda_v \frac{k^\rho}{n_\text{cal}^\tau}\| v \|_\delta + \lambda_M \frac{(k(k\!-\!1))^\rho}{n_\text{cal}^\tau}\| M \|_\delta \; .
\end{multline}
%We choose not to regularize $\alpha$, assuming that linear scaling does not overfit.
We apply penalties on the norm of each parameter group: the diagonal $v$, the off-diagonal $(\mathbf{1}_k \mathbf{1}_k^\top - I_k) \odot M$ and the intercept $b$.
Intuitively, the number of parameters in each group (respectively $k$, $k(k-1)$ and $k$) as well as the number of calibration samples $n_\text{cal}$ should have an effect on the regularization strength.
Thus, we weight each penalty by $n_\text{cal}^{-\tau}$ and the group size to the power $\rho$.
Finally, each group penalty has a specific weight $\lambda_b$, $\lambda_v$ and $\lambda_M$. The hyperparameters of our regularization scheme are thus:
\begin{itemize}[topsep=0pt, itemsep=0pt]
    \item The order $\delta$ of the norm used.
    \item The exponents $\tau, \rho$ used for the number of samples and the parameter group size.
    \item The group-specific weights $\lambda_v$, $\lambda_M$ and $\lambda_b$.
\end{itemize}

\cref{eq:FittingSMS} is a softmax regression on the logits with a hierarchical regularization structure.
We can also simplify by restricting our hierarchical model to a vector-scaling (SVS) model
$$
 g_\text{SVS}(x) = S \big( ( I_k + \text{diag}(v)) S^{-1}(x) + b \big) \, ,
$$
which we fit by solving
\begin{equation}\label{eq:FittingSVS}
    \min_{b, v} \frac{1}{n_\text{cal}} \sum_{i=1}^{n_\text{cal}} \ell(g_\text{SVS}(x_i), y_i) + \lambda_b \frac{k^\rho}{n_\text{cal}^\tau}\|b\|_\delta
    + \lambda_v \frac{k^\rho}{n_\text{cal}^\tau}\| v \|_\delta .
\end{equation}
Since we first apply temperature scaling, regularization in \eqref{eq:FittingSMS} and \eqref{eq:FittingSVS} can be used with the same parameters for more or less confident classifiers, overcoming a central limitation of pre-existing regularized post hoc calibration methods.
It remains to select good values for the regularization parameters that we introduced.
In \cref{appendix:HyperparameterSearch} we describe how we derive a unique set of regularization parameters that deliver robust calibration performance across different sample sizes and problem dimensions for both SMS and SVS.
The following section presents experiment results for SVS and SMS using these default parameters.

\section{EXPERIMENTS}
\label{sec:MulticlassExperiments}

\subsection{Implementation}
We release Python solvers to fit vector \eqref{eq:FittingSVS} and matrix scaling \eqref{eq:FittingSMS} with our hierarchical regularization structure, for user-specified regularization hyperparameters.
We implement a L-BFGS based solver \cite{liu1989limited} using the pytorch-minimize package \citep{feinman2021pytorch}, inspired by the implementations from the \texttt{torchcal} package by \cite{ranjan2023torchcal}.
Such quasi-Newton solvers however assume differentiability and thus only support ridge regularization.

To support composite objectives with non-smooth regularizers, thereby enabling logistic post-hoc calibration with a richer family of penalties, we also implement a solver that uses the SAGA algorithm \citep{defazio2014saga}.
We optimize execution time for this solver with just-in-time compilation using \texttt{Numba} \citep{lam2015numba}.

We make these new calibration functions available in scikit-learn classifier style modules: \texttt{SVSCalibrator} and \texttt{SMSCalibrator} in our open source package \texttt{probmetrics}.
Usage is illustrated in \cref{code:MulticlassSnippet}.

\begin{lstlisting}[language=Python, label=code:MulticlassSnippet, caption=Post-hoc calibration with SMS., numbers=none]
from probmetrics.calibrators import SMSCalibrator

p_cal = my_model.predict_proba(X_cal)

sms = SMSCalibrator(
    penalty="ridge", rho=1.0, tau=1.0,
    lambda_intercept=1.0,
    lambda_diagonal=1.0,
    lambda_off_diagonal=1.0,
    opt="saga"
)
sms.fit(p_cal, y_cal)

# Initial preds:
p_test = my_model.predict_proba(X_test)
# Calibrated preds:
p_test = sms.predict_proba(p_test)
\end{lstlisting}

In our modules, exponents $\rho$ and $\tau$ and group-specific coefficients $\lambda_b, \lambda_v$ and $\lambda_M$ can be set manually. The penalty used (which corresponds to the order of the norm $\delta$) can be one of: MCP \citep{zhang2010nearly}, LASSO \citep{tibshirani1996regression}, group LASSO \citep{yuan2006model} (using the SAGA solver) and ridge (using either SAGA or L-BFGS).

As described in the previous section, the impact of regularization in \eqref{eq:FittingSMS} and \eqref{eq:FittingSVS} depends on the scale of the logits. For smaller logits, larger changes in parameter value are needed to achieve the same effect.
To make our configuration of the regularization parameters robust to more or less confident classifiers, we apply SVS and SMS on logits rescaled using temperature scaling.
In practice, we pre-process probabilities with the temperature scaling implementation presented in  \citet{berta2025rethinking} (using Laplace smoothing to avoid infinite temperatures when calibration accuracy is 100\%) before fitting SVS or SMS.
% This allows us to fix the global scaling parameter $\alpha$ to one, yielding faster convergence.

\begin{remark}
    Logistic calibration functions take as input probabilities predicted by the initial model on the calibration set but operate on the logits of theses predicted probabilities.
    In practice, computing logits from predicted probabilities needs to be implemented carefully.
    In our package, we first compute the log probabilities (that can contain infinite values) and then clip these to the log of the smallest normal float32 (around -90).
\end{remark}

\subsection{Tabular Experiments}
\label{subsec:TabularExperiments}

\begin{table}[htbp]
\centering
\resizebox{\linewidth}{!}{
\begin{tabular}{l|l|c|c}
\toprule
Type & Method & Brier score diff. & Logloss diff. \\
\midrule
Non param. & Isotonic regression & -0.0026 & 0.1893 \\
\midrule
\multirow{3}{4em}{TS} & Guo et al. & 0.0112 & 0.0278 \\
& torchcal & -0.0024 & -0.0116 \\
& probmetrics & \textbf{-0.0024} & \textbf{-0.0148} \\
\midrule
\multirow{2}{4em}{VS} & torchcal & -0.0032 & -0.0053 \\
& SVS & \textbf{-0.0034} & \textbf{-0.0174} \\
\midrule
\multirow{3}{4em}{MS} & torchcal & 0.0018 & 0.0642 \\
& Dirichlet & -0.0023 & -0.0089 \\
& SMS & \textbf{-0.0046} & \textbf{-0.0210} \\
\bottomrule
\end{tabular}
}
\caption{\emph{Average absolute differences in test Brier score and logloss after recalibration (lower is better) for SVS and SMS, compared with other calibration methods.}
The average is taken over our $1365$ multi-class experiments.}
\label{tab:MulticlassAbsoluteResults}
\end{table}

\paragraph{Benchmark data.}
To benchmark the different multi-class calibration methods that we introduced, we use \texttt{TabRepo}, a large set of model predictions provided by \citet{salinas2023tabrepo}.
\texttt{TabRepo} stores predictions obtained by training classical machine and deep learning models on dozens of tabular datasets for a variety of hyperparameter configurations.
We use the predictions of $M = 7$ different models: logistic regression, CatBoost \citep{dorogush2018catboost}, LightGBM \citep{ke2017lightgbm}, XGBoost \citep{chen2016xgboost}, random forest \citep{breiman2001random}, 
and two neural nets from FastAI \citep{howard2020fastai} and AutoGluon \citep{erickson_autogluon-tabular_2020}.
For these models, predictions are available for $D = 65$ multi-class classification datasets.
For each dataset, the data is split in ten folds, and $T = 3$ separate tasks are created by using one of the first three folds as a test set.
For each task, the remaining data is used for eight-fold cross-validation, and the predictions of the eight models on their respective validation folds are concatenated and provided as a validation set, which we use to fit post-hoc calibration functions.
This results in a total of $M \times D \times T = 1365$ experiments for our benchmark.

Models in \texttt{TabRepo} are fitted using the training data and early stopping is performed based the validation logloss (except for linear regression and random forest, for which no early stopping is required).
For every experiment, predictions are available for the model trained with default hyperparameters as well as for 200 randomly generated hyperparameter configurations.
We evaluate our post-hoc calibration functions on the models trained with the hyperparameter configuration that achieves the smallest validation error.
As explained above, we fit post-hoc calibration functions using the validation data  and report the gains obtained on the test set.

\begin{figure}
    \centering
    \includegraphics[width=\linewidth]{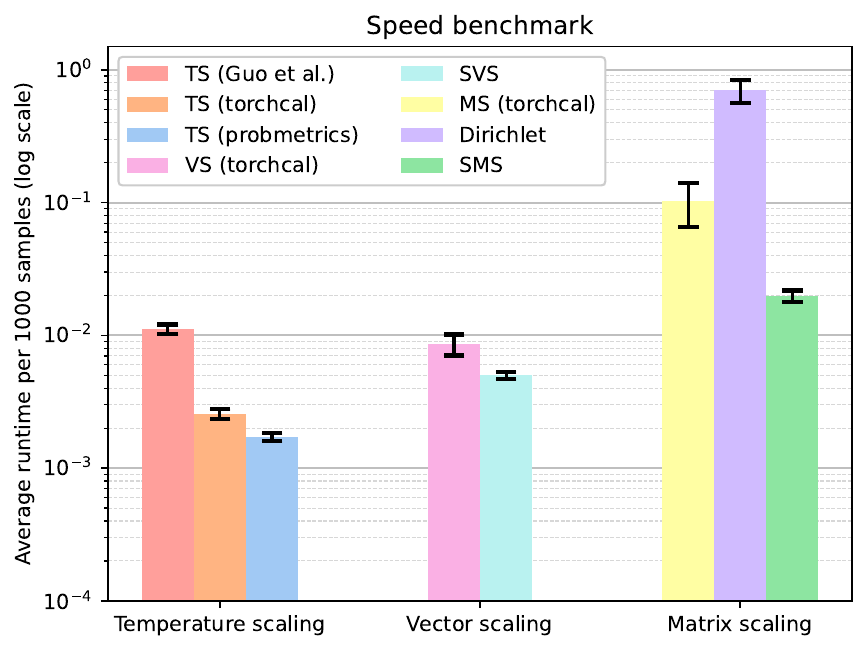}
    \caption{\emph{Average calibration function fitting time per 1000 samples.} We compute the average using all our 1365 multi-class experiments. Error bars are standard $95\%$ confidence intervals, treating each model-pair as independent (455 experiments). Average runtimes (y-axis) are plotted using a log scale.}
    \label{fig:MulticlassBenchmarkTime}
\end{figure}

\paragraph{Methods.} We compare SVS and SMS (with default regularization hyperparameters) against logistic recalibration baselines.
As a general baseline, we use standard (non-regularized, as introduced by \cite{guo2017calibration}) implementations of TS, VS and MS available in the \texttt{torchcal} package by \cite{ranjan2023torchcal}. It uses pytorch-minimize \citep{feinman2021pytorch} to minimize the logloss after post-hoc calibration on the calibration set.
For TS, we also report the original implementation by \citet{guo2017calibration}, and the implementation previously presented in the \texttt{probmetrics} package \citep{berta2025rethinking}.
Finally, for matrix scaling, we report results obtained with Dirichlet calibration \citep{kull2019dirichlet}. In the original Dirichlet calibration paper, the authors recommend performing a grid search to select the best regularization parameters on every dataset, using cross-validation on the calibration set. Given that fitting Dirichlet calibration once on our 1365 experiments is already very slow (\Cref{fig:MulticlassBenchmarkTime}), we fix the regularization parameters to $10^{-3}$, the default value recommended in the supplementary material.
Note that Dirichlet calibration can be recovered with a certain choice of hyperparameters in our matrix scaling framework \eqref{eq:FittingSMS}.
As we will see, our experiments reveal that this choice is suboptimal.
Finally, we also include one-versus-rest isotonic regression \citep{zadrozny2002transforming} in our benchmark, which is arguably the most standard non-parametric baseline.

\paragraph{Metrics.} We compare different calibration methods in terms of test logloss and Brier score after recalibration.
Logloss and Brier score being proper scoring rules \citep{gneiting2007strictly}, they evaluate forecast quality by measuring calibration error plus a second term called refinement error \citep{brocker2009reliability, kull2015novel}.
However, since refinement error cannot be improved by post-hoc calibration (see for example corollary A.3 in \citep{berta2025rethinking}), observed differences in test loss after recalibration can only come from variation of the calibration error (or increased refinement error, which should be penalized).
For this reason, the difference between test loss before and after calibration is a well grounded way to compare post-hoc calibration methods, and does not require on a top-label approximation in the multi-class case.

\begin{figure}
    \centering
    \includegraphics[width=\linewidth]{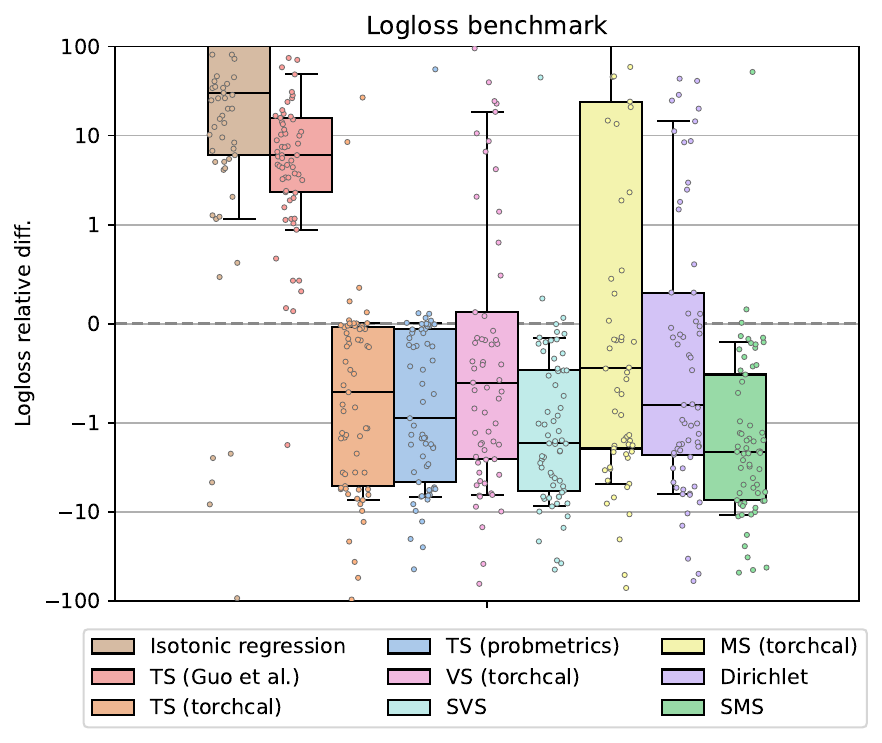}
    \caption{\emph{Relative differences in test logloss after recalibration (lower is better) for SVS and SMS, compared with other calibration methods.}
    Each dot represents the average relative loss difference obtained for one tabular dataset, over 21 experiments (7 models, 3 folds).
    Box-plots show the 10, 25, 50, 75, and 90\% quantiles. Relative differences (y-axis) are plotted using a log scale and clipped to -100\% loss (min) and +100\% loss (max).}
    \label{fig:MulticlassRelativeResults}
\end{figure}

\paragraph{Results.}
In \cref{tab:MulticlassAbsoluteResults} we report \emph{absolute} test loss differences after calibration (lower is better), averaged over our 1365 experiments.
In \cref{fig:MulticlassRelativeResults} we plot \emph{relative} test logloss differences after calibration on a scale from $+100\%$ to $-100\%$, the same plot for Brier score is in \cref{appendix:AdditionalMulticlassResults}.

The TS implementation by \cite{guo2017calibration} has a convergence issue and thus performs poorly\footnote{After we reported it to the authors, this issue was fixed.
The implementation now performs better than reported.}.
Because it can assign zero probability to some classes, which sometimes leads to infinite logloss, isotonic regression is very bad for test logloss.
In terms of Brier score however, it is competitive with well-implemented TS but not with SVS or SMS.

Logistic baselines suffer from overfitting: for \texttt{torchcal} implementations, more parameters translate to more datasets with increased logloss after recalibration.
Matrix scaling drastically overfits with degraded test performance on almost half the datasets.
Even with regularization, Dirichlet only partially controls overfitting and underperforms standard temperature or vector scaling (while being orders of magnitude slower to run).

Our hierarchical regularization structure, on the other hand, is effective at preventing overfitting and allows for consistent improvement as the number of parameters increases.
Using insights from \cref{sec:multiclassGaussianModel}, we know that more complex recalibration functions alleviate the constraints on the logit distribution of the initial model~$f$.
In some scenarios, temperature scaling might be far from well-specified while vector and matrix scaling can help to get closer to the optimal recalibration function $g^*$, which explains the observed improvement.
On \cref{fig:MulticlassRelativeResults}, we see that even for large quantiles of the loss difference distribution, SVS and SMS do not overfit. The gains reported on the bottom of the distributions are, however, growing with the number of parameters.
This demonstrates that we successfully built a post-hoc calibration method that navigates the tradeoff between under- and overfitting the calibration set, by adapting to the number of classes~$k$ and calibration points $n_\text{cal}$.
In \cref{appendix:AdditionalMulticlassResults}, we plot relative differences separately for each model in our benchmark.

\paragraph{Effect of number of samples and number of classes.}
In \cref{appendix:AdditionalMulticlassResults} we group our $65$ multi-class datasets by growing number of calibration samples and number of classes, to study the effect on the different post-hoc calibration methods compared.
We see that while the performance of non-regularized techniques like matrix scaling varies widely with the number of classes and calibration samples available, our regularized functions perform well across every group.
In particular, SMS ranks best in average, whatever the number of classes and calibration samples.

\paragraph{Statistical analysis.}
The results presented so far do not allow us to conclude formally that SVS and SMS outperform existing methods with statistical significance.
The impact of post-hoc calibration has a large variability across models and dataset, making it hard to rely on confidence intervals to compare methods.
To ensure a rigorous statistical comparison, we follow the standard protocol for benchmarking algorithms over multiple datasets established by \cite{demvsar2006statistical}.
Instead of treating all experiments as independent (which would violate independence assumptions due to correlations between folds and models on the same data), we aggregate loss differences at the dataset level ($D = 65$).

We first employ the Friedman test to detect statistical differences across methods, followed by the Nemenyi post-hoc test to identify specific pairwise differences.
This non-parametric approach is the standard in machine learning evaluation as it is robust to non-normal distributions of performance metrics and corrects for multiple comparisons, ensuring that the reported improvements are not due to chance or inflated degrees of freedom.

\begin{figure}
    \centering
    \includegraphics[width=\linewidth]{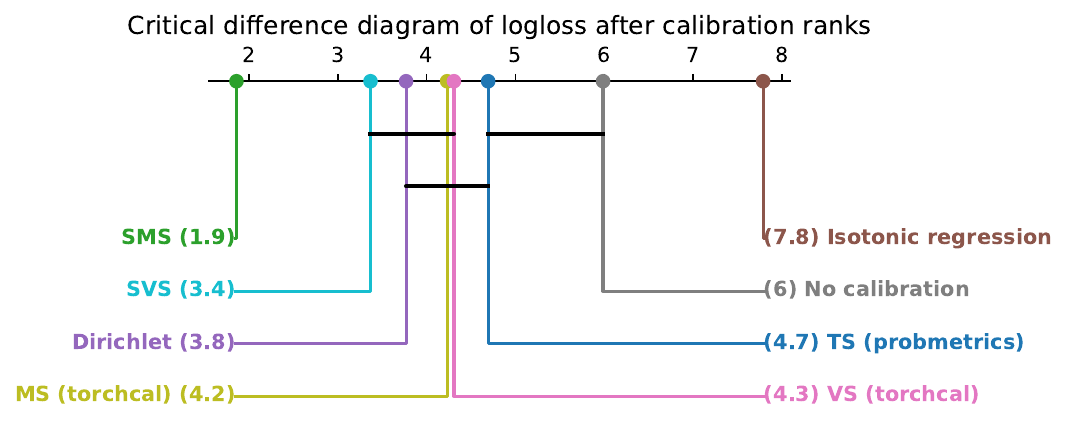}
    \caption{\emph{Critical difference diagram for logloss after calibration.}
    Black lines indicate statistically indistinguishable methods. Numbers in parentheses indicate the average rank of each method on our 65 datasets (lower is better).}
    \label{fig:MulticlassCD}
\end{figure}

In \cref{fig:MulticlassCD}, we plot the results of the statistical analysis using a critical difference diagram.
The diagram shows the average rank of each method (lower is better).
Black horizontal lines connecting methods indicate ``statistical indistinguishability''.
We add the “No calibration” method to the benchmark (average loss difference $= 0$ for every dataset).
The critical diagram obtained for Brier score is reported in \cref{appendix:AdditionalMulticlassResults}.

The key takeaway is that on our tabular benchmark, SMS is the sole winner, as it is the only method that beats every other, with statistical significance.
SVS is statistically tied with Dirichlet, vector scaling and matrix scaling.
Only TS is statistically indistinguishable from no calibration, while isotonic regression is the worst method for logloss (for Brier score it outperforms only no-calibration).

\paragraph{Computational time.}
On \cref{fig:MulticlassBenchmarkTime}, we report the average fitting time per 1000 samples of the different methods compared.
All experiments were run locally, on CPU, on a MacBook Pro equipped with an Apple M2 Pro chip.
Our implementations are faster than \texttt{torchcal} on average, despite the fact that they include regularization.
When compared to Dirichlet calibration, SMS is roughly 70 times faster.
Given that it is standard practice to select regularization parameters with a grid search and cross-validation \citep{kull2019dirichlet}, or to perform post-hoc calibration repeatedly, when estimating refinement error for early stopping \citep{berta2025rethinking}, these time considerations can make large differences for users.

Our speed benchmark was performed using the default regularization parameters and the L-BFGS solver for SVS and SMS.
For the same configuration, the SAGA solver is slightly slower and needs to compile when called for the first time (it uses \texttt{Numba} for just-in-time compilation), which takes a few seconds.
However it is much more flexible since it allows non-convex penalties.

All experiments and figures can be reproduced using our experimental repository, at \url{https://github.com/eugeneberta/LogisticCalibrationBenchmark}.

\subsection{Computer Vision Experiments}

To benchmark our recalibration functions on computer vision experiments, we use logits for several deep neural network architectures trained on CIFAR-10, CIFAR-100 and ImageNet, provided by \citet{kull2019dirichlet}.

In \cref{tab:LoglossCIFAR10}, \cref{tab:LoglossCIFAR100} and \cref{tab:LoglossImageNet} we report the absolute difference in logloss after recalibration with \texttt{probmetrics}' temperature scaling (TS probm.), \texttt{torchcal}'s vector and matrix scaling, Dirichlet calibration, SVS and SMS, for CIFAR-10, CIFAR-100 and ImageNet respectively.
For ImageNet, there are 1000 classes, so fitting matrix scaling requires more that $10^6$ parameters which is prohibitively large, hence we only report results for temperature and vector scaling.

We see that the largest test loss improvement is always achieved by SVS or SMS, with SMS providing the best recalibration in all cases but one. Sometimes, large improvements over temperature scaling and baselines can be achieved.
For CIFAR-100, which has a large number of classes, we see that non-regularized matrix scaling (torchcal MS) has skyrocketing test loss and Dirichlet scaling's regularization fails to prevent overfitting.
While SVS seems only marginally better than non-regularized vector scaling on CIFAR-10 and CIFAR-100, we see on the ImageNet results that regularization makes a difference when the number of classes becomes very large, which could become even more striking for unbalanced datasets.

\begin{table}[htbp]
\centering
\resizebox{\linewidth}{!}{
\begin{tabular}{l|c|cc|ccc}
\toprule
 & TS & \multicolumn{2}{c|}{VS} & \multicolumn{3}{c}{MS} \\
Model & probm. & SVS & torchcal & SMS & torchcal & Dirichlet \\
\midrule
densenet40 & -0.206 & -0.208 & -0.206 & \textbf{-0.209} & -0.203 & -0.206 \\
lenet5 & -0.023 & -0.073 & -0.076 & \textbf{-0.080} & -0.077 & -0.076 \\
resnet110 & -0.149 & -0.151 & -0.150 & \textbf{-0.155} & -0.146 & -0.149 \\
resnet110-SD & -0.128 & -0.130 & -0.128 & \textbf{-0.130} & -0.116 & -0.121 \\
resnet-wide32 & -0.191 & -0.199 & -0.199 & \textbf{-0.201} & -0.189 & -0.192 \\
\bottomrule
\end{tabular}}
\caption{Logloss absolute improvement on CIFAR-10.}
\label{tab:LoglossCIFAR10}
\end{table}

\begin{table}[htbp]
\centering
\resizebox{\linewidth}{!}{
\begin{tabular}{l|c|cc|ccc}
\toprule
 & TS & \multicolumn{2}{c|}{VS} & \multicolumn{3}{c}{MS} \\
Model & probm. & SVS & torchcal & SMS & torchcal & Dirichlet \\
\midrule
densenet40 & -0.961 & -0.967 & -0.958 & \textbf{-0.970} & 28.982 & 1.075 \\
lenet5 & -0.134 & -0.219 & -0.266 & \textbf{-0.273} & 8.204 & 0.166 \\
resnet110 & -0.602 & \textbf{-0.608} & -0.598 & -0.608 & 35.361 & 0.400 \\
resnet110-SD & -0.410 & -0.423 & -0.421 & \textbf{-0.437} & 37.586 & 1.201 \\
resnet-wide32 & -0.858 & -0.864 & -0.851 & \textbf{-0.869} & 25.516 & 0.540 \\
\bottomrule
\end{tabular}}
\caption{Logloss absolute improvement on CIFAR-100.}
\label{tab:LoglossCIFAR100}
\end{table}

\begin{table}[htbp]
\centering
\small
\begin{tabular}{l|c|cc}
\toprule
 & TS & \multicolumn{2}{c}{VS} \\
Model & probm. & SVS & torchcal VS \\
\midrule
densenet161 & -0.035 & \textbf{-0.037} & 0.007 \\
resnet152 & -0.047 & \textbf{-0.049} & -0.015 \\
\bottomrule
\end{tabular}
\caption{Logloss absolute improvement on ImageNet.}
\label{tab:LoglossImageNet}
\end{table}

\section{CONCLUSION}
We have demonstrated that parametric post-hoc calibration functions based on logistic regression can be motivated theoretically for both binary and multi-class classification. This insight naturally leads to the use of more expressive calibration methods beyond standard temperature scaling, such as matrix scaling in the multi-class setting.  To turn such methods into a practical tool, it is necessary to address the fundamental tradeoffs that arise when models of different complexity are used for the post-hoc calibration functions.  We have shown how such tradeoffs can be addressed with novel regularization schemes. Through extensive experiments, we demonstrate that such schemes can yield substantial improvements over existing logistic-based post-hoc calibration techniques.

Our implementations, released in an open-source github package, offer significant performance gains and computational efficiency compared to existing methods.
They accommodate various penalties, including MCP, LASSO, group LASSO, and ridge, with independent tuning of regularization strengths across parameter groups, offering the user the opportunity to find the best regularization parameters for every scenario.
For strong out-of-the-box performance, we propose default hyperparameters that generalize well across varying numbers of classes and calibration points, and leverage meta-learning to validate this default configuration.

The empirical results consistently show that our approach yields better recalibration, with more complex models outperforming simpler ones. Our regularization scheme effectively mitigates overfitting, allowing more expressive models to capture complex miscalibration patterns without degrading generalization performance.
This work highlights the importance of carefully designed regularization in unlocking the full potential of richer calibration functions, making them a compelling replacement for common temperature, vector, and matrix scaling implementations.

\subsubsection*{Acknowledgements}
The authors would like to thank Adrien Taylor for discussions regarding this work.

This publication is part of the Chair ``Markets and Learning'', supported by Air Liquide, BNP PARIBAS ASSET MANAGEMENT Europe, EDF, Orange and SNCF, sponsors of the Inria Foundation.

This work received support from the French government, managed by the National Research Agency, under the France 2030 program with the reference ``PR[AI]RIE-PSAI'' (ANR-23-IACL-0008).

Funded by the European Union (ERC-2022-SYG-OCEAN-101071601). Views and opinions expressed are however those of the author(s) only and do not necessarily reflect those of the European Union or the European Research Council Executive Agency. Neither the European Union nor the granting authority can be held responsible for them.

\bibliography{references}

\clearpage
\appendix
\thispagestyle{empty}

% Supplementary material: To improve readability, you must use a single-column format for the supplementary material.
\onecolumn
\aistatstitle{Structured Matrix Scaling for Multi-class Calibration: \\
Supplementary Materials}

\numberwithin{figure}{section}

\section{BINARY LOGISTIC CALIBRATION}
\label{appendix:BinaryCase}

In the binary classification setting, we observe some random variable $X \in \mathcal{X}$ and try to predict the value of some other (binary) random variable $Y \in \{+1,-1\}$, traditionally known as the \textit{outcome}. Statistical models typically make continuous predictions $p \in [0,1]$, where larger $p$ indicates larger certainty from the model that the outcome is $Y=+1$. In the following we denote $\sigma$ the sigmoid function $\sigma(x) = (1+\exp(-x))^{-1}$ and $\sigma^{-1}$ its inverse, the logit function $\sigma^{-1}(x) = \log \frac{x}{1-x}$. $\mathbb{S}_+^{d \times d}$ is the space of $d \times d$ positive semidefinite matrices.

\subsection{Motivating Calibration with Logistic Regression} \label{sec:binaryGaussianModel}

Many existing binary parametric post-hoc calibration methods derive from logistic regression on the log odds $\sigma^{-1}(p) = \log \frac{p}{1-p}$ of the predicted probabilities $f(X) = p$. We will see that this choice can be motivated from a simple setting.

Assume $\mathcal{X} = \mathbb{R}^d$, and consider a binary class-conditional Gaussian data model $\mathbb{P}(X\,|\,Y=+1) = \mathcal{N}(\mathbf{\mu_+}, \Sigma_+)$ and $\mathbb{P}(X\,|\,Y=-1) = \mathcal{N}(\mathbf{\mu_-}, \Sigma_-)$ with mean vectors $\mu_+, \mu_- \in \mathbb{R}^d$, and covariance matrices $\Sigma_+, \Sigma_- \in \mathbb{S}_+^{d \times d}$. We denote the class probabilities $\pi_+ = \mathbb{P}(Y=+1)$ and $\pi_- = \mathbb{P}(Y=-1)$. Let $f$ be a logistic regression classifier parametrized by some weight vector $w \in \mathbb{R}^d$: $f(X) = \sigma(w^\top X)$.
We have that the logits $w^\top X$ conditioned on each class follow a one-dimensional Gaussian distribution:
$$
\mathbb{P}(w^\top X \,|\, Y = +1) = \mathcal{N}(w^\top \mathbf{\mu_+}, w^\top \Sigma_+ w) \, , \quad \mathbb{P}(w^\top X \,|\, Y = -1) = \mathcal{N}(w^\top \mathbf{\mu_-}, w^\top \Sigma_- w) \, .
$$ 

Denoting the means $m_+ := w^\top \mu_+, m_- := w^\top \mu_-$ and variances $\sigma_+^2 := w^\top \Sigma_+ w, \sigma_-^2 := w^\top \Sigma_- w$ we have that
$$
    \mathbb{P}(Y=+1\,|\, w^\top X = x) = \sigma( ax^2 + bx + c ) \, ,
$$
with
$$
a = \frac{1}{2 \sigma_-^2} - \frac{1}{2 \sigma_+^2} \, , \quad
b = \frac{m_+}{\sigma_+^2} - \frac{m_-}{\sigma_-^2} \, , \quad
c = \log\Big(\frac{\pi_+\sigma_-}{\pi_-\sigma_+}\Big) + \frac{m_-^2}{2\sigma_-^2} - \frac{m_+^2}{2\sigma_+^2} \, .
$$
See \cref{appendix:Proofs} for a proof.
We want our classifier $f$ to be calibrated, that is
$$
    \mathbb{P}(Y=+1 \,|\, f(X) = s) = s \, .
$$
In our example, $f(X) = s \iff \sigma(w^\top X) = s \iff w^\top X = \sigma^{-1}(s)$. So the distribution of $Y$ is
$$
\mathbb{P}(Y=+1 \,|\, f(X)=s) = \sigma \big( a \sigma^{-1}(s)^2 + b \sigma^{-1}(s) + c \big) \, .
$$
Denoting $g$ the post-hoc calibration function $g(s) = \sigma \big( a \sigma^{-1}(s)^2 + b \sigma^{-1}(s) + c \big)$, it holds that $\mathbb{P}(Y=+1 \,|\, f(X)=s) = g(s)$. In general, the function $g$ does not have to be an injection; for a given value $g \circ f = p$ we define the set $S_p = \{ s \,|\, g(s) = p \}$, then
\begin{align*}
    \mathbb{P}(Y=+1 \,|\, g \circ f(X) = p) &= \int_{S_p} \mathbb{P}(Y=+1 \,|\, f(X) = s) \mathbb{P}(f(X) = s \,|\, g \circ f(X) = p) ds\\
    &= \int_{S_p} p \mathbb{P}(f(X) = s \,|\, g \circ f(X) = p) ds\\
    &= p \, ,
\end{align*}
which proves that $g \circ f$ is calibrated.

\begin{remark}
    In our motivating example, the logistic calibration function is well specified because the class-conditional logit distributions are Gaussian. However, performing post-hoc calibration with logistic regression does not correspond to assuming that the class-conditional distributions are Gaussian as it is sometimes suggested in the literature. There is a wide range of distribution pairs for which the conditional is logistic, as discussed in \cite{jordan1995logistic} and \cite{boken2021appropriateness}. It holds as soon as logistic regression is well specified, see \citep{bishop2006pattern}.
\end{remark}

\subsection{Logistic Calibration Functions}

Our example setting in \cref{sec:binaryGaussianModel} motivated the use of a logistic model as a post-hoc calibration function.
In practice however, we cannot compute coefficients $a$, $b$ and $c$ directly.
For a real world dataset, the data is not normally distributed and our classifier does not have to be logistic regression so logistic recalibration might not be well specified.
We hope there exists some parameters of $g$ for which $g \circ f$ is approximately calibrated.
We resort to empirical risk minimization to find such coefficients.
Using the reserved calibration set $(x_i, y_i)_{1 \leq i \leq n_\text{cal}}$, we solve
\begin{equation} \label{eq:binaryLogisticRegression}
    \min_g \frac{1}{n_\text{cal}} \sum_{i=1}^{n_\text{cal}} \ell(g \circ f(x_i), y_i) \, ,
\end{equation}
where $g$ is parametrized by the coefficients of the logistic model. $\ell$ is typically the logloss, which makes the problem convex and thus allows us to find the optimal coefficients easily.

In the literature, different choices for the function $g$ based on the logistic model coexist:
\begin{itemize}
    \item Linear scaling (known as temperature scaling \citep{guo2017calibration}): $g(x) = \sigma( \alpha \sigma^{-1}(x))$, $g$ is parametrized only by a linear coefficient $\alpha$.
    \item Affine scaling (known as Beta scaling \citep{kull2017beta}): $g(x) = \sigma(\alpha \sigma^{-1}(x) + \beta)$, $g$ is parametrized by both a linear and constant coefficient $\alpha$ and $\beta$.
    \item To the best of our knowledge, quadratic scaling: $g(x) = \sigma(\gamma \sigma^{-1}(x)^2 + \alpha \sigma^{-1}(x) + \beta)$, has not been used for post-hoc calibration before. We include it in our experiments.
\end{itemize}

In the binary case, we only have 1, 2 or 3 parameters (respectively for linear, affine and quadratic scaling) to learn, so the risk of overfitting the empirical risk estimate is limited. We evaluate our three methods without regularization, although our implementation can accommodate different penalties on any of the three parameters.
Before moving to empirical results, we look back at our example from \cref{sec:binaryGaussianModel} to get some insight about the different design choices for $g$.

\paragraph{The affine hypothesis.}
Looking back at our data model, the affine hypothesis corresponds to setting $a=0$, which implies $\sigma_- = \sigma_+$. In other words, affine scaling is well specified only when the two class-wise logit distributions have the same variance.
We have no guarantee that this should hold in practice, suggesting that quadratic scaling might bring further improvement over affine scaling in scenarios where this assumption is not satisfied.

\paragraph{The linear hypothesis.}
The linear hypothesis is even more constraining. Setting $c=0$ puts constraints on the means, variances, but also class probabilities $\pi_+$ and $\pi_-$. We see no reason why this should be true in practice and we thus expect to see a large performance gap between linear and affine scaling.

\subsection{Experiments} \label{sec:BinaryExperiments}

\subsubsection{Implementation}
For linear, affine and quadratic scaling, we provide efficient Python implementations that solve the convex optimization problem \eqref{eq:binaryLogisticRegression}.
We use the default scikit-learn implementation for logistic regression, that uses L-BFGS \cite{liu1989limited} and we remove regularization.
Depending on wether we want to apply linear, affine or quadratic scaling, we feed only logits $(\sigma^{-1}(p_i))_{1 \leq i \leq n_\text{cal}}$, a constant and logits $(1, \sigma^{-1}(p_i))_{1 \leq i \leq n_\text{cal}}$ or a constant, logits and squared logits $(1, \sigma^{-1}(p_i), \sigma^{-1}(p_i)^2)_{1 \leq i \leq n_\text{cal}}$ to the classifier.
We release this into a calibration module \texttt{BinaryLogisticCalibrator} in our open-source package \texttt{probmetrics}.
For ease of use, we follow the scikit-learn classifier format as illustrated in \cref{code:BinarySnippet}.

\begin{lstlisting}[language=Python, label=code:BinarySnippet, caption=Post-hoc calibration with probmetric's binary calibrator., numbers=none]
from probmetrics.calibrators import BinaryLogisticCalibrator

p_cal = my_model.predict_proba(X_cal)

quad_scaling = BinaryLogisticCalibrator(type="quadratic")
quad_scaling.fit(p_cal, y_val)

p_test = my_model.predict_proba(X_test) # Initial preds
p_test = quad_scaling.predict_proba(p_test) # Calibrated preds
\end{lstlisting}

We also release SAGA-based implementations for affine and quadratic scaling \citep{defazio2014saga}, with execution time optimized using just-in-time compilation with \texttt{Numba} \citep{lam2015numba}.
Since SAGA can accommodate non smooth penalties, these are compatible with: MCP \citep{zhang2010nearly}, LASSO \citep{tibshirani1996regression} and ridge, should the user want to apply regularization on the intercept and/or quadratic parameters for affine scaling and quadratic scaling.
We refer to our package for details on the implementation, regularization options and running examples.

\begin{remark}
    Logistic calibration functions take as input probabilities predicted by the initial model on the calibration set but operate on the logits of theses predicted probabilities.
    The binary logit function that maps predicted probabilities to logits $\sigma^{-1}(p) = \log \frac{p}{1-p}$ needs to be implemented carefully in practice.
    In our package, to compute binary logits, we first compute $\log(p) - \log(1-p)$ (using the numerically stable \texttt{log1p} function). This potentially returns infinite values, so we clip the outcome to minus and plus the log of the smallest normal float32 (around $\pm90$).
\end{remark}

\subsubsection{Tabular Experiments}

\paragraph{Data.}
To benchmark the different binary calibration methods that we introduced, we use \texttt{TabRepo}, a large set of model predictions provided by \citet{salinas2023tabrepo}.
\texttt{TabRepo} stores predictions obtained by training classical machine and deep learning models on dozens of tabular datasets for a variety of hyperparameter configurations.
We use the predictions of $M = 7$ different models: logistic regression, CatBoost \citep{dorogush2018catboost}, LightGBM \citep{ke2017lightgbm}, XGBoost \citep{chen2016xgboost}, random forest \citep{breiman2001random}, 
and two neural nets from FastAI \citep{howard2020fastai} and AutoGluon \citep{erickson_autogluon-tabular_2020}.
For these models, predictions are available for $D = 104$ multi-class classification datasets.
For each dataset, the data is split in ten folds, and $T = 3$ separate tasks are created by using one of the first three folds as a test set.
For each task, the remaining data is used for eight-fold cross-validation, and the predictions of the eight models on their respective validation folds are concatenated and provided as a validation set, which we use to fit post-hoc calibration functions.
This results in a total of $M \times D \times T = 2184$ experiments for our tabular benchmark.

Models in \texttt{TabRepo} are fitted using the training data and early stopping is performed based the validation logloss (except for linear regression and random forest, for which no early stopping is required).
For every experiment, predictions are available for the model trained with default hyperparameters as well as for 200 randomly generated hyperparameter configurations.
We evaluate our post-hoc calibration functions on the models trained with the hyperparameter configuration that achieves the smallest validation error.
As explained above, we fit post-hoc calibration functions using the validation data  and report the gains obtained on the test set.

\begin{remark}
    For data-efficiency, we reuse the validation set to fit the post-hoc calibration function. Note that it is already used for early stopping (except for logistic regression and random forest), and to select the best hyperparameter configuration. We acknowledge this deviation from standard protocol could risk overfitting. However, our empirical results on the test set demonstrate substantial improvements from recalibration, suggesting that generalization was not compromised.
\end{remark}

\paragraph{Methods.}
We compare linear scaling with existing temperature scaling implementations: the original implementation by \citet{guo2017calibration}, and an implementation previously presented in the \texttt{probmetrics} package \citep{berta2025rethinking}.
For affine and quadratic scaling, we report the performance of our implementations.

\begin{figure}
    \centering
    \includegraphics[width=0.5\linewidth]{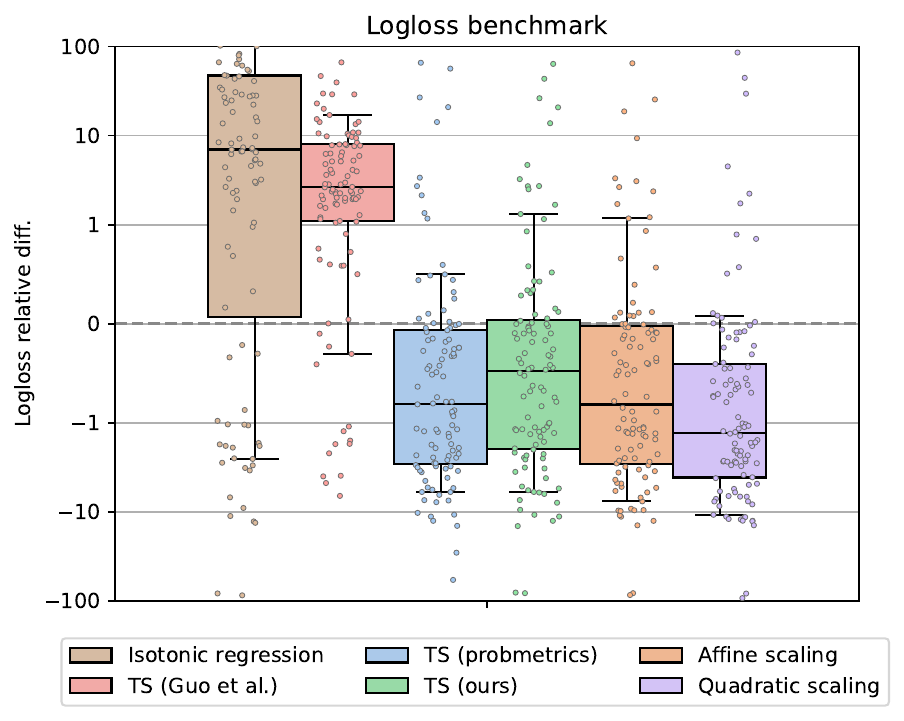}
    \caption{\emph{Relative differences in test logloss (lower is better) after recalibration for our linear, affine and quadratic scaling functions, compared with other existing implementations.}
    Each dot represents the average relative loss difference obtained for one tabular dataset, over 21 experiments (7 models, 3 folds).
    Box-plots show the 10, 25, 50, 75, and 90\% quantiles. Relative differences (y-axis) are plotted using a log scale and clipped to -100\% loss (min) and +100\% loss (max).}
    \label{fig:BinaryBenchmarkLogloss}
\end{figure}

\paragraph{Results.}
In \cref{tab:BinaryAbsoluteResults} we report \emph{absolute} test loss differences after calibration (lower is better), averaged over our 2184 experiments.
In \cref{fig:BinaryBenchmarkLogloss} we plot \emph{relative} test logloss differences after calibration on a scale from $+100\%$ to $-100\%$, the same plot for Brier score is in \cref{appendix:AdditionalBinaryResults}.

\begin{table}
\centering
\small
\begin{tabular}{l|c|c}
\toprule
Method & Brier score gap & Logloss gap \\
\midrule
Isotonic regression & -0.0025 & 0.0909 \\
TS (probmetrics) & -0.0018 & -0.0167 \\
TS (Guo et al.) & 0.0063 & 0.0043 \\
TS (ours) & -0.0018 & -0.0142 \\
Affine scaling & -0.0026 & -0.0154 \\
Quadratic scaling & \textbf{-0.0034} & \textbf{-0.0181} \\
\bottomrule
\end{tabular}
\caption{\emph{Average absolute differences in test Brier score and logloss after recalibration (lower is better) for binary calibration methods.}
The average is taken over our $2184$ binary experiments.}\label{tab:BinaryAbsoluteResults}
\end{table}

The TS implementation by \cite{guo2017calibration} has a convergence issue and thus performs poorly\footnote{After we reported it to the authors, this issue was fixed.
The implementation now performs better than reported.}.
Because it can assign zero probability to some classes, which sometimes leads to infinite logloss, isotonic regression is very bad for test logloss.
In terms of Brier score however, it is competitive with well-implemented TS and even affine scaling but seems to under-perform quadratic scaling.

The difference between \texttt{probmetrics}' TS and ours comes from the fact that \texttt{probmetrics} uses Laplace smoothing on the predicted probabilities, which has a positive impact for the test logloss.
As expected, affine scaling yields a noticeable improvement over standard temperature scaling.
Interestingly, we observe important additional gains for quadratic scaling over affine scaling.
In average, on our benchmark, this new method performs best.

In \cref{appendix:AdditionalBinaryResults}, we plot improvements separately for each model in our benchmark.

\paragraph{Effect of number of samples and number of classes.}
In \cref{appendix:AdditionalBinaryResults} we group our $104$ binary datasets by growing number of calibration samples, to study the effect on the different post-hoc calibration methods compared.
We see that while linear, affine and quadratic scaling perform similarly when the number of calibration samples is small, quadratic scaling becomes better than the two other (less parametrized) methods with more calibration samples.

\paragraph{Statistical analysis.}
The results presented so far do not allow us to conclude formally that quadratic scaling outperforms existing methods with statistical significance.
The impact of post-hoc calibration has a large variability across models and dataset, making it hard to rely on confidence intervals to compare methods.
To ensure a rigorous statistical comparison, we follow the standard protocol for benchmarking algorithms over multiple datasets established by \cite{demvsar2006statistical}.
Instead of treating all experiments as independent (which would violate independence assumptions due to correlations between folds and models on the same data), we aggregate loss differences at the dataset level ($D = 104$).

We first employ the Friedman test to detect statistical differences across methods, followed by the Nemenyi post-hoc test to identify specific pairwise differences.
To do so, we use the \texttt{scikit-posthocs} Python package \citep{terpilowski2019scikit}.
This non-parametric approach is the standard in machine learning evaluation as it is robust to non-normal distributions of performance metrics and corrects for multiple comparisons, ensuring that the reported improvements are not due to chance or inflated degrees of freedom.

\begin{figure}
    \centering
    \includegraphics[width=0.6\linewidth]{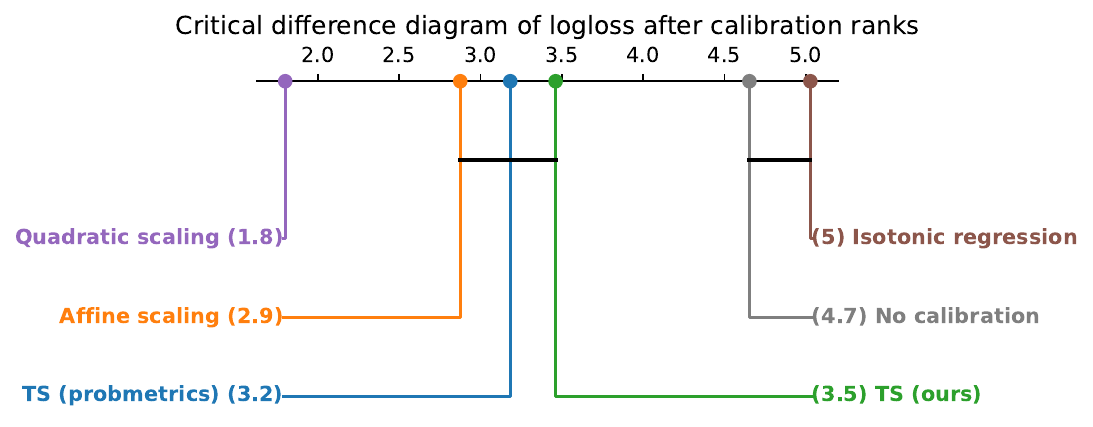}
    \caption{\emph{Critical differences diagram for logloss after calibration.}
    Black lines indicate statistically indistinguishable methods. Number in parentheses indicate the average rank of each method on the 104 datasets (lower is better).}
    \label{fig:BinaryCDLogloss}
\end{figure}

In \cref{fig:BinaryCDLogloss}, we plot the results of the statistical analysis using a critical difference diagram.
The diagram shows the average rank of each method (lower is better).
Black horizontal lines connecting methods indicate ``statistical indistinguishability''.
We add the “No calibration” method to the benchmark (average loss difference $= 0$ for every dataset).
The critical diagram obtained for brier score is reported in \cref{appendix:AdditionalMulticlassResults}.

The key takeaway is that on our tabular benchmark, quadratic scaling is the sole winner, as it is the only method that beats every other, with statistical significance.
Affine scaling is statistically tied with temperature scaling.
Only isotonic regression is statistically indistinguishable from no calibration (for logloss, it performs better for Brier score).

\begin{figure}
    \centering
    \includegraphics[width=0.5\linewidth]{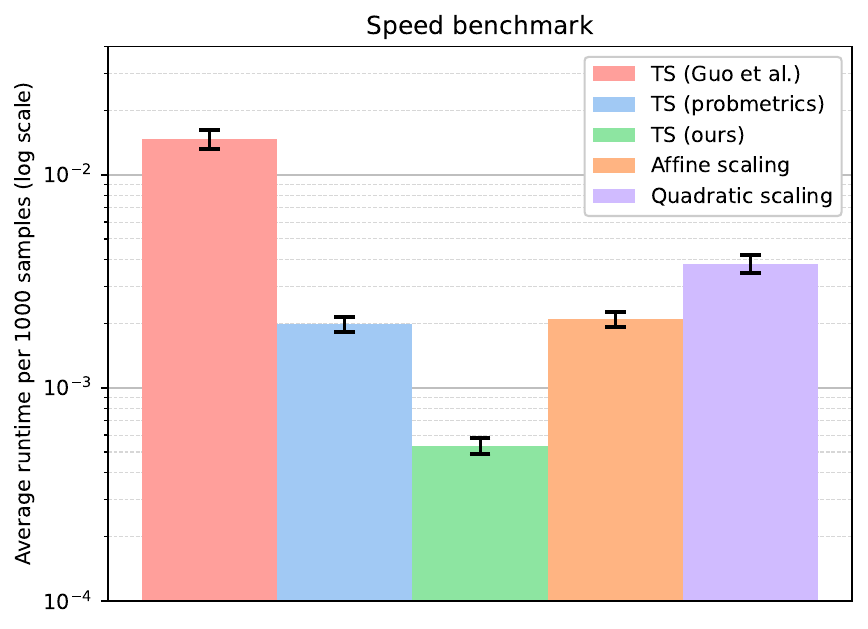}
    \caption{\emph{Average calibration function fitting time per 1000 samples.} We compute the average using all our 2184 binary experiments. Error bars are standard $95\%$ confidence intervals, treating each model-pair as independent (728 experiments). Average runtimes (y-axis) are plotted using a log scale.}
    \label{fig:BinaryBenchmarkTime}
\end{figure}

\paragraph{Computational time.}
On \cref{fig:BinaryBenchmarkTime} we plot the average fitting time per 1000 samples for the different binary calibration functions in our benchmark.
All experiments were run locally, on CPU, on a MacBook Pro equipped with an Apple M2 Pro chip.
Our implementation is the fastest available for temperature scaling.
Affine and quadratic scaling are slower but still almost as fast as \texttt{probmetrics}' TS.
These runtime considerations can be crucial when post-hoc calibration is performed repeatedly, for example when estimating refinement error for early stopping \citep{berta2025rethinking}.

All experiments and figures can be reproduced using our experimental repository, at \url{https://github.com/eugeneberta/LogisticCalibrationBenchmark}.

\newpage
\section{PROOFS}
\label{appendix:Proofs}

\paragraph{Proof in the Binary Case.}
Denoting $m_+ := w^\top \mu_+, m_- := w^\top \mu_-$ and $\sigma_+^2 := w^\top \Sigma_+ w, \sigma_-^2 := w^\top \Sigma_- w$,
$$
\mathbb{P}(w^\top X = x \,|\, Y = +1) = \mathcal{N}(x \,|\, m_+, \sigma^2_+) \, , \quad \mathbb{P}(w^\top X = x \,|\, Y = -1) = \mathcal{N}(x \,|\, m_-, \sigma^2_-) \, .
$$
Let us prove that $\mathbb{P}(Y=+1\,|\, w^\top X = x) = \sigma( ax^2 + bx + c )$, with
$$
a = \frac{1}{2 \sigma_-^2} - \frac{1}{2 \sigma_+^2} \, , \quad
b = \frac{m_+}{\sigma_+^2} - \frac{m_-}{\sigma_-^2} \, , \quad
c = \log\Big(\frac{\pi_+\sigma_-}{\pi_-\sigma_+}\Big) + \frac{m_-^2}{2\sigma_-^2} - \frac{m_+^2}{2\sigma_+^2} \, .
$$
Using Bayes' theorem,
\begin{align*}
    \mathbb{P}(Y=+1\,|\,w^\top X = x) &= \frac{\mathbb{P}(w^\top X = x\,|\,Y=+1) \mathbb{P}(Y=+1)}{\mathbb{P}(w^\top X = x)}\\
    &= \frac{\mathbb{P}(w^\top X = x\,|\, Y=+1) \mathbb{P}(Y=+1)}{\mathbb{P}(w^\top X = x\,|\,Y=+1) \mathbb{P}(Y=+1) + \mathbb{P}(w^\top X = x\,|\,Y=-1) \mathbb{P}(Y=-1)}\\
    &= \frac{\mathcal{N}(x \,|\, m_+, \sigma_+^2) \pi_+}{\mathcal{N}(x \,|\, m_+, \sigma_+^2) \pi_+ + \mathcal{N}(x \,|\, m_-, \sigma_-^2) \pi_-}\\
    &= \frac{1}{1+\frac{\pi_-}{\pi_+} \frac{\mathcal{N}(x \,|\, m_-, \sigma_-^2)}{\mathcal{N}(x \,|\, m_+, \sigma_+^2)}}\\
    &= \frac{1}{1 + \frac{\pi_- \sigma_+}{\pi_+ \sigma_-} \exp \Big( - \frac{(x-m_-)^2}{2\sigma_-^2} + \frac{(x-m_+)^2}{2\sigma_+^2} \Big) }\\
    &= \sigma \Big( \frac{(x-m_-)^2}{2\sigma_-^2} - \frac{(x-m_+)^2}{2\sigma_+^2} + \log \frac{\pi_+ \sigma_-}{\pi_- \sigma_+}  \Big) \\
    &= \sigma \Big( (\frac{1}{2 \sigma_-^2} - \frac{1}{2 \sigma_+^2}) x^2 + (\frac{m_+}{\sigma_+^2} - \frac{m_-}{\sigma_-^2})x + (\frac{m_-^2}{2 \sigma_-^2} - \frac{m_+^2}{2 \sigma_+^2} + \log \frac{\pi_+ \sigma_-}{\pi_- \sigma_+} ) \Big)\\
    &= \sigma( ax^2 + bx + c ) \, ,
\end{align*}
which concludes the proof.

\paragraph{Proof in the Multi-class Case.}

Denoting $m_i := C_k W \mu_i$, $\sigma_i := C_k W \Sigma_i W^\top C_k^\top$ and $\mathbb{P}(Y=i) = \pi_i$ for all $i \in \{ 0, \cdots, k \} $, we have
$$
\mathbb{P}(C_kWX = x\,|\,Y=i) = \frac{ \exp \big( -\frac{1}{2} (x - m_i)^\top \sigma_i^+ (x - m_i) \big)}{\sqrt{\,|\,2 \pi \sigma_i\,|\,_+}} \, .
$$
Let us prove that
$$
    \mathbb{P}(Y \,|\, C_k WX = x) =  S(x^\top \mathbf{A} x + B x + C) \, ,
$$
with $\mathbf{A}$ a tensor in $\mathbb{R}^{k \times k \times k}$ defined by $\mathbf{A}[i,:,:] = - \frac{1}{2} \sigma_i^+$ so that $(x^\top \mathbf{A} x)_i = - \frac{1}{2} x^T \sigma_i^+ x$, $B$ a matrix in $\mathbb{R}^{k \times k}$ defined by $B_i = m_i^\top \sigma_i^+$ and $C$ a vector in $\mathbb{R}^k$ defined by $C_i = - \frac{1}{2} m_i^\top \sigma_i^+ m_i + \log(\pi_i \,|\,\sigma_i\,|\,_+^{-1/2})$.
Using Bayes' theorem,
\begin{align*}
    \mathbb{P}(Y=i \,|\, C_k WX = x) &= \frac{\mathbb{P}(C_k WX = x \,|\, Y=i) \mathbb{P}(Y=i)}{\mathbb{P}(C_k WX = x)}\\
    &= \frac{\mathbb{P}(C_k WX = x \,|\, Y=i) \mathbb{P}(Y=i)}{\sum_{j=1}^k \mathbb{P}(C_kWX = x \,|\, Y=j) \mathbb{P}(Y=j)}\\
    &= \frac{ \pi_i \,|\,\sigma_i\,|\,_+^{-1/2} \exp \big( -\frac{1}{2} (x-m_i)^\top \sigma_i^+ (x-m_i) \big) }{\sum_{j=1}^k \pi_j \,|\,\sigma_j\,|\,_+^{-1/2} \exp \big( -\frac{1}{2} (x-m_j)^\top \sigma_j^+ (x-m_j) \big)}\\
    &= \frac{ \exp \big( -\frac{1}{2} (x-m_i)^\top \sigma_i^+ (x-m_i) + \log(\pi_i \,|\,\sigma_i\,|\,_+^{-1/2}) \big) }{\sum_{j=1}^k \exp \big( -\frac{1}{2} (x-m_j)^\top \sigma_j^+ (x-m_j) + \log(\pi_j \,|\,\sigma_j\,|\,_+^{-1/2}) \big)}\\
    &= S(x^\top \mathbf{A} x + B x + C)_i \, ,
\end{align*}
which concludes the proof.

\section{HYPERPARAMETER SEARCH}
\label{appendix:HyperparameterSearch}

For both SVS and SMS, we aim to find good values for the regularization parameters $\delta, \rho, \tau, \lambda_b, \lambda_v$, and $\lambda_M$.
Ideally, we would like to derive a single parameter configuration that makes our algorithm robust to overfitting, while leaving room for additional complexity when enough data is available and recalibration can be improved further.
Our goal is that with well-chosen parameters, our methods should outperform non-regularized linear, vector and matrix scaling, demonstrating that the correct regularization level can be selected adaptively.
As a baseline, we consider setting all hyperparameters to one: $\lambda_b = \lambda_v = \lambda_M = 1$, $\tau = \rho = 1$.

To try to improve beyond this default choice, we employ meta-learning \citep{vanschoren2018meta}.
Using model predictions from \texttt{TabRepo} \citep{salinas2023tabrepo}, we compare the performance of $g_\text{SVS}$ and $g_\text{SMS}$ for different regularization parameter configurations.
In \texttt{TabRepo}, predictions are stored for $D = 65$ multi-class classification datasets.
We use the predictions of $M = 7$ different models.
For each dataset, the data is split in ten folds, and $T = 3$ separate tasks are created by using one of the first three folds as a test set.
This results in a total of $M \times D \times T = 1365$ experiments.
See \cref{subsec:TabularExperiments} for details on the data used.

We first perform temperature scaling on all experiments and rank the datasets by the average test Brier score improvement obtained.
We split the datasets into two groups by picking, in this sorted list, odd-indexed datasets for our hyperparameter search and even-indexed datasets to form a control group.
This process ensures that the potential calibration gains are well distributed between the datasets we use for the parameter search and the control group.

We then evaluate the average performance of $g_\text{SVS}$ and $g_\text{SMS}$ on the $33$ hyperparameter search datasets, using the following grid of hyperparameters:
\begin{itemize}[topsep=0pt, itemsep=0pt, leftmargin=10pt]
    \item The order $\delta$ of the norm used for regularization can be one of: $\ell_1$ (LASSO regularization \citep{tibshirani1996regression}) which promotes sparsity on individual parameters, $\ell_2$ (group LASSO \citep{yuan2006model}) which promotes sparsity at the group level—setting the whole off-diagonal group to zero for example—and $\ell_2^2$ (classical ridge regression). After experimenting with different values for $\delta$, we observe more stable  results with ridge regression, so we used this penalty for our hyperparameter search.
    \item Intuitively, groups with more parameters are more susceptible to exhibit overfitting, and we thus expect that number of parameters should have a positive impact on the penalty applied to the group.
    For example, off-diagonal parameters should be more regularized than diagonal parameters.
    \citet{kull2019dirichlet}, on the other hand, heuristically employ an off-diagonal regularization scheme that is inversely proportional to the number of parameters. For completeness, we explore both positive and negative values for the exponent of the parameter group size $\rho$, and we consider the grid $\{ -1, -0.5, 0., 0.5, 1 \}$.
    \item Fewer regularization samples translates to higher risk of overfitting, so we only explore positive values for the exponent of the number of samples $\tau$: $\{ 0.5, 1, 1.5, 2 \}$.
    \item For the group specific coefficients $\lambda_b$, $\lambda_v$ and $\lambda_M$, we explore continuous values in the range $[0.01, 100]$ using a logarithmic scale.
\end{itemize}

We use Optuna \citep{optuna_2019} to optimize the hyperparameter configuration based on the average test Brier score improvement obtained on all hyperparameter search datasets ($693$ experiments per configuration in total).
We refer the interested reader to our experimental repository\footnote{\url{https://github.com/eugeneberta/LogisticCalibrationBenchmark}} for details on the hyperparameter search.
As expected, hyperparameter configurations with positive values for~$\rho$ generally perform better than when negative values are used, contradicting the design choice of \citet{kull2019dirichlet}.

We then compare the results obtained with the best hyperparameter configurations for SVS and SMS on the $32$ control datasets (remaining $672$ experiments) with our baseline hyperparameter choice $\lambda_b = \lambda_v = \lambda_M = 1$, $\tau = \rho = 1$. While the tuned configurations outperform the baseline on the hyperparameter search datasets, we observe similar recalibration performance on the control datasets.
With default parameters, the optimization problems reduce to
$$
\min_{\alpha, b, v} \frac{1}{n_\text{cal}} \sum_{i=1}^{n_\text{cal}} \ell(g_\text{SVS}(x_i), y_i) + \frac{k}{n_\text{cal}} \|b\|_2^2 + \frac{k}{n_\text{cal}} \| v \|_2^2
$$
for SVS and
\[
    \min_{\alpha, b, v, M} \frac{1}{n_\text{cal}} \sum_{i=1}^{n_\text{cal}} \ell(g_\text{SMS} (x_i), y_i) + \frac{k}{n_\text{cal}}\|b\|_2^2 + \frac{k}{n_\text{cal}}\| v \|_2^2 + \frac{k(k - 1)}{n_\text{cal}}\| M \|_2^2
\]
for SMS.

In our multi-class experiments (\cref{sec:MulticlassExperiments}), we use these default hyperparameters for simplicity, and we report results for the entire set of $65$ multi-class datasets ($1365$ experiments).

\section{ADDITIONAL BINARY RESULTS}
\label{appendix:AdditionalBinaryResults}

\begin{figure}[htbp]
    \centering
    \includegraphics[width=0.5\linewidth]{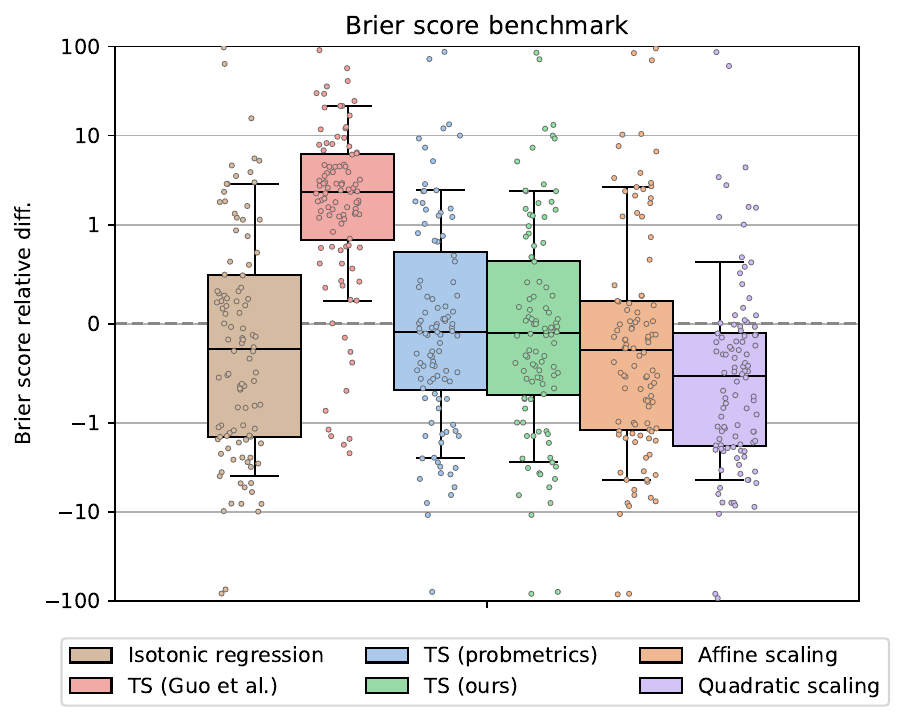}
    \caption{\emph{Relative differences in test Brier score after recalibration (lower is better) for our linear, affine and quadratic scaling functions, compared with other existing implementations.}
    Each dot represents the average relative loss difference obtained for one tabular dataset, over 21 experiments (7 models, 3 folds).
    Box-plots show the 10, 25, 50, 75, and 90\% quantiles. Relative differences (y-axis) are plotted using a log scale and clipped to -100\% loss (min) and +100\% loss (max).}
\end{figure}

\begin{figure}[htbp]
    \centering
    \includegraphics[width=0.7\linewidth]{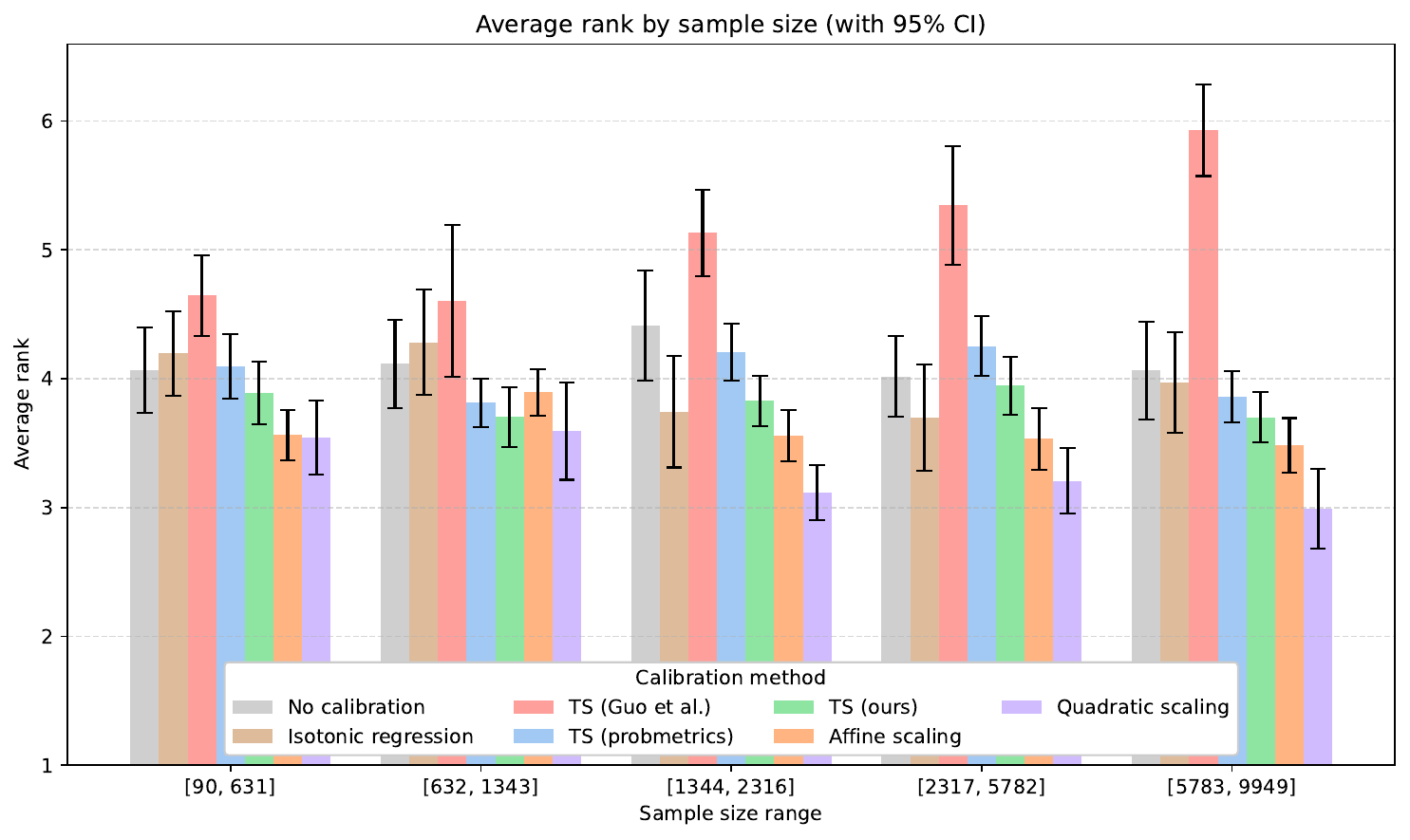}
    \caption{\emph{Average rank of binary calibration methods, per number of calibration samples.}
    Datasets are grouped by growing number of calibration samples, using equal mass binning with 5 bins.
    Each bin contains $104/5 \simeq 21$ datasets. 
    On each bin, we rank calibration methods by average test Brier score improvement on the dataset.
    We plot standard $95\%$ confidence intervals for the average ranks.}
\end{figure}

\begin{figure}[htbp]
    \centering
    \includegraphics[width=0.9\linewidth]{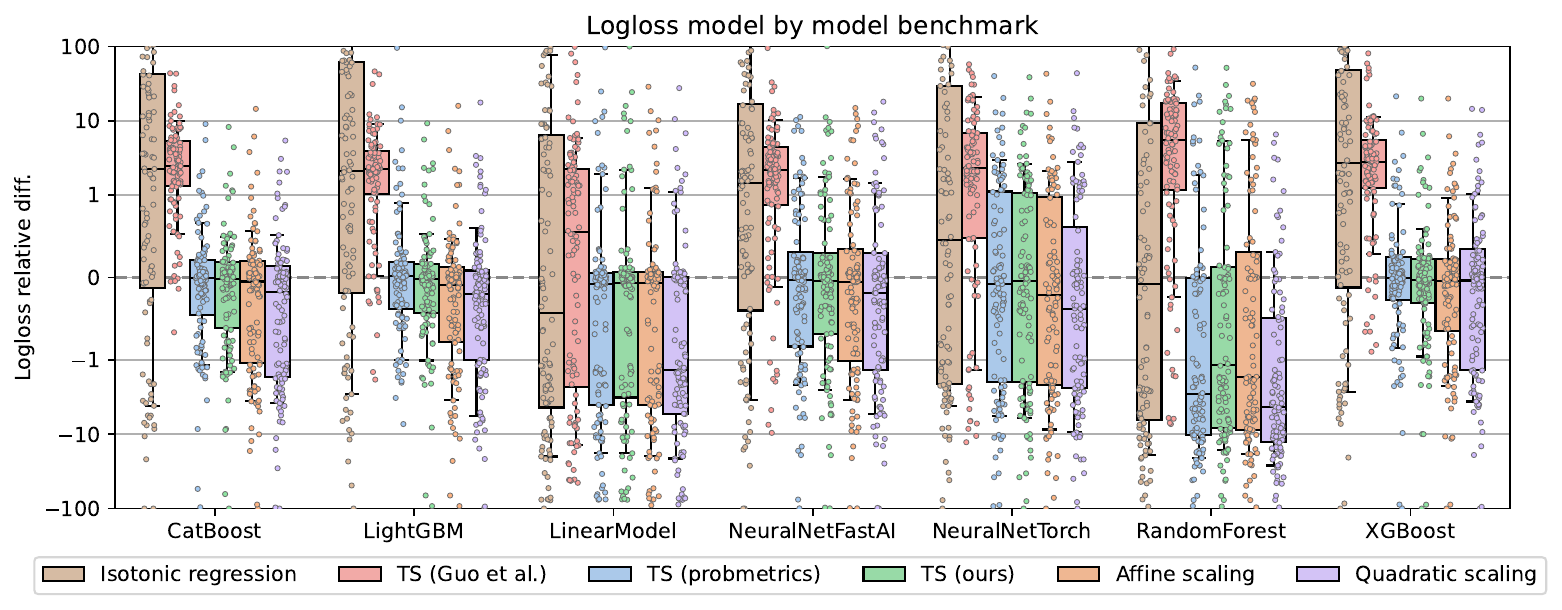}
    \caption{\emph{Relative differences in test logloss (lower is better) after recalibration for our linear, affine and quadratic scaling functions, compared with other existing implementations, separately for every model in our benchmark.}
    Each dot represents the average relative loss difference obtained for one model on one tabular dataset, over 3 folds.
    Box-plots show the 10, 25, 50, 75, and 90\% quantiles. Relative differences (y-axis) are plotted using a log scale and clipped to -100\% loss (min) and +100\% loss (max).}
\end{figure}

\begin{figure}[htbp]
    \centering
    \includegraphics[width=0.9\linewidth]{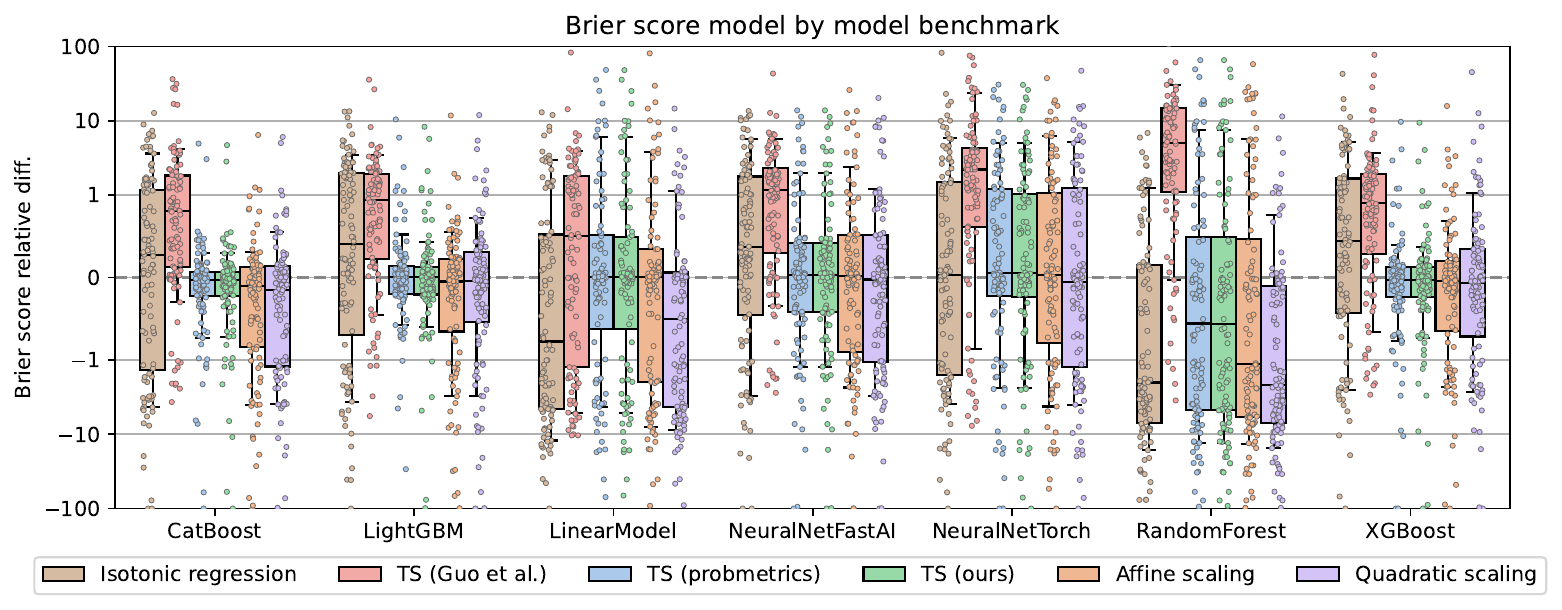}
    \caption{\emph{Relative differences in test Brier score (lower is better) after recalibration for our linear, affine and quadratic scaling functions, compared with other existing implementations, separately for every model in our benchmark.}
    Each dot represents the average relative score difference obtained for one model on one tabular dataset, over 3 folds.
    Box-plots show the 10, 25, 50, 75, and 90\% quantiles. Relative differences (y-axis) are plotted using a log scale and clipped to -100\% loss (min) and +100\% loss (max).}
\end{figure}

\begin{figure}
    \centering
    \includegraphics[width=0.6\linewidth]{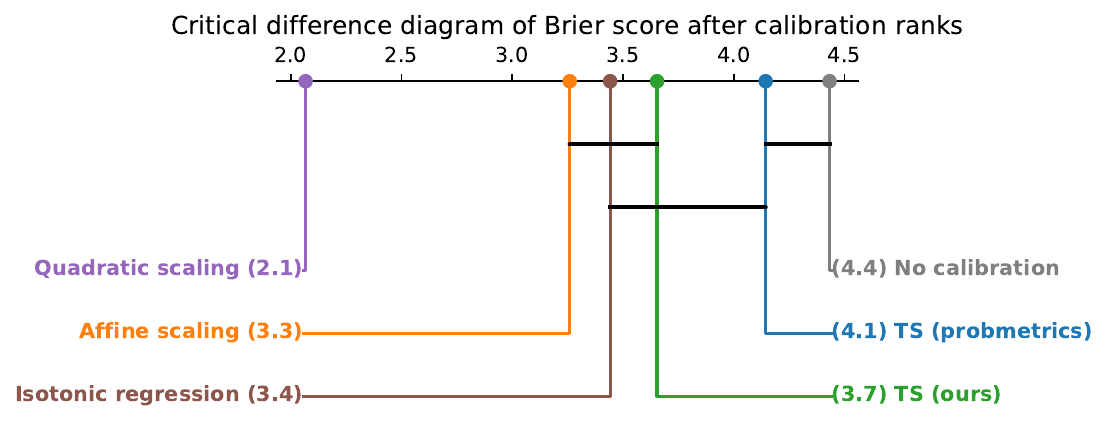}
    \caption{\emph{Critical differences diagram for Brier score after calibration.}
    Black lines indicate statistically indistinguishable methods. Number in parentheses indicate the average rank of each method on the 104 datasets (lower is better).}
    \label{fig:BinaryCDBrier}
\end{figure}

\section{ADDITIONAL MULTI-CLASS RESULTS}
\label{appendix:AdditionalMulticlassResults}

\begin{figure}[htbp]
    \centering
    \includegraphics[width=0.5\linewidth]{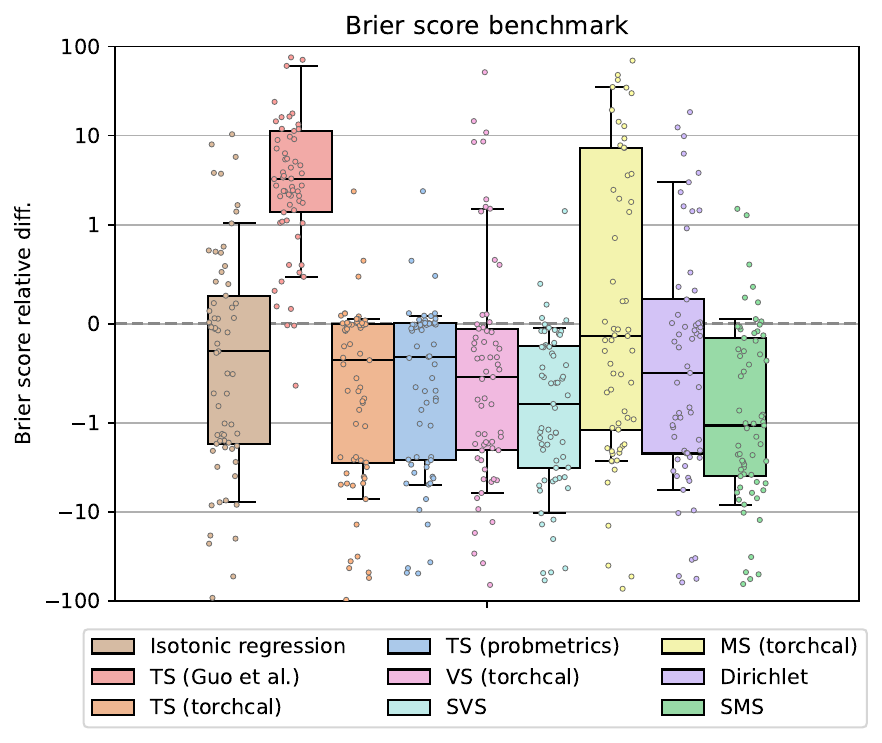}
    \caption{\emph{Relative differences in test Brier score after recalibration (lower is better) for SVS and SMS, compared with other calibration methods.}
    Each dot represents the average relative loss difference obtained for one tabular dataset, over 21 experiments (7 models, 3 folds).
    Box-plots show the 10, 25, 50, 75, and 90\% quantiles. Relative differences (y-axis) are plotted using a log scale and clipped to -100\% loss (min) and +100\% loss (max).}
\end{figure}

\begin{figure}[htbp]
    \centering
    \includegraphics[width=0.7\linewidth]{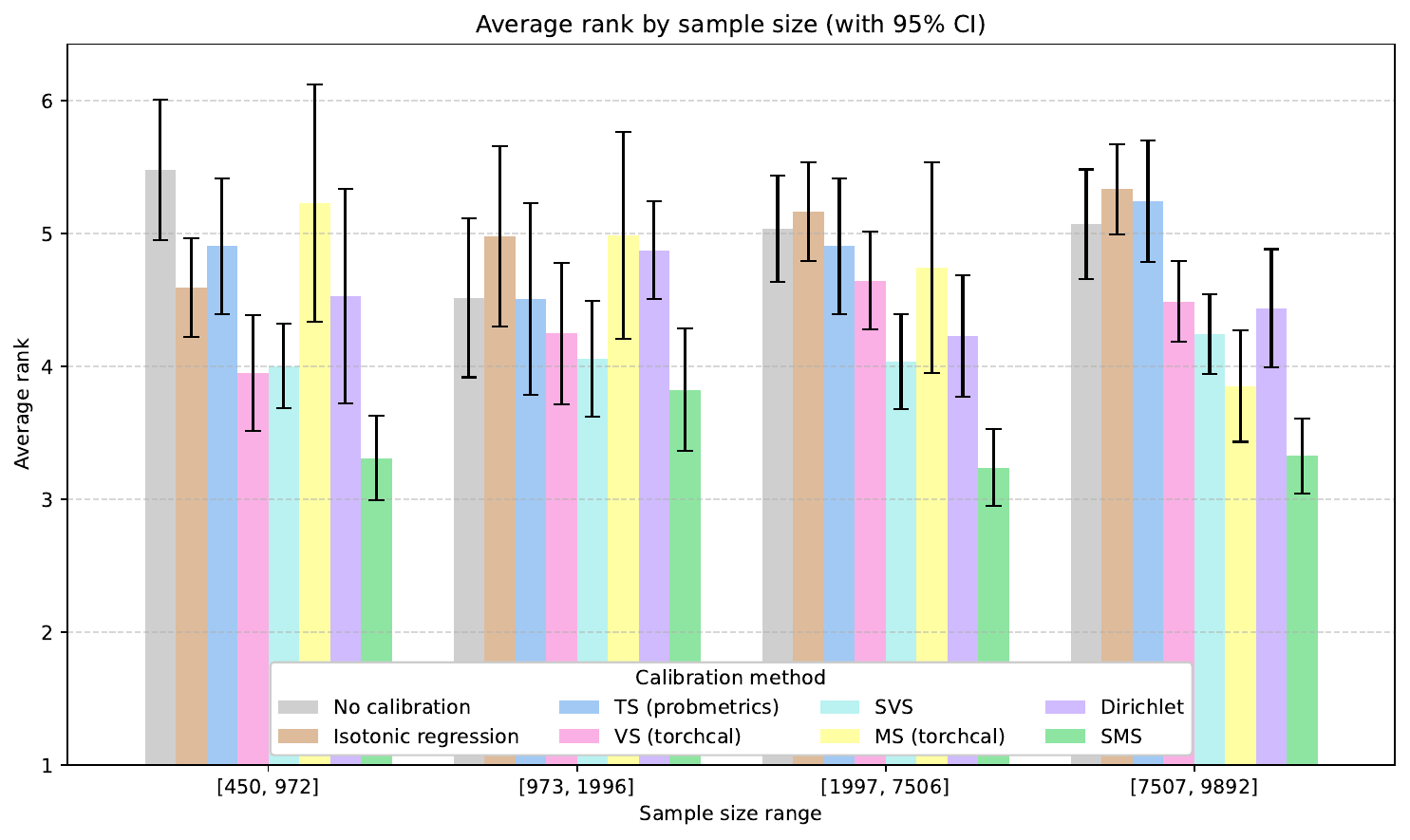}
    \caption{\emph{Average rank of multi-class calibration methods, per number of calibration samples.}
    Datasets are grouped by growing number of calibration samples, using equal mass binning with 4 bins.
    Each bin contains $65/4 \simeq 16$ datasets. 
    On each bin, we rank calibration methods by average test Brier score improvement on the dataset.
    We plot standard $95\%$ confidence intervals for the average ranks.}
\end{figure}

\begin{figure}[htbp]
    \centering
    \includegraphics[width=0.7\linewidth]{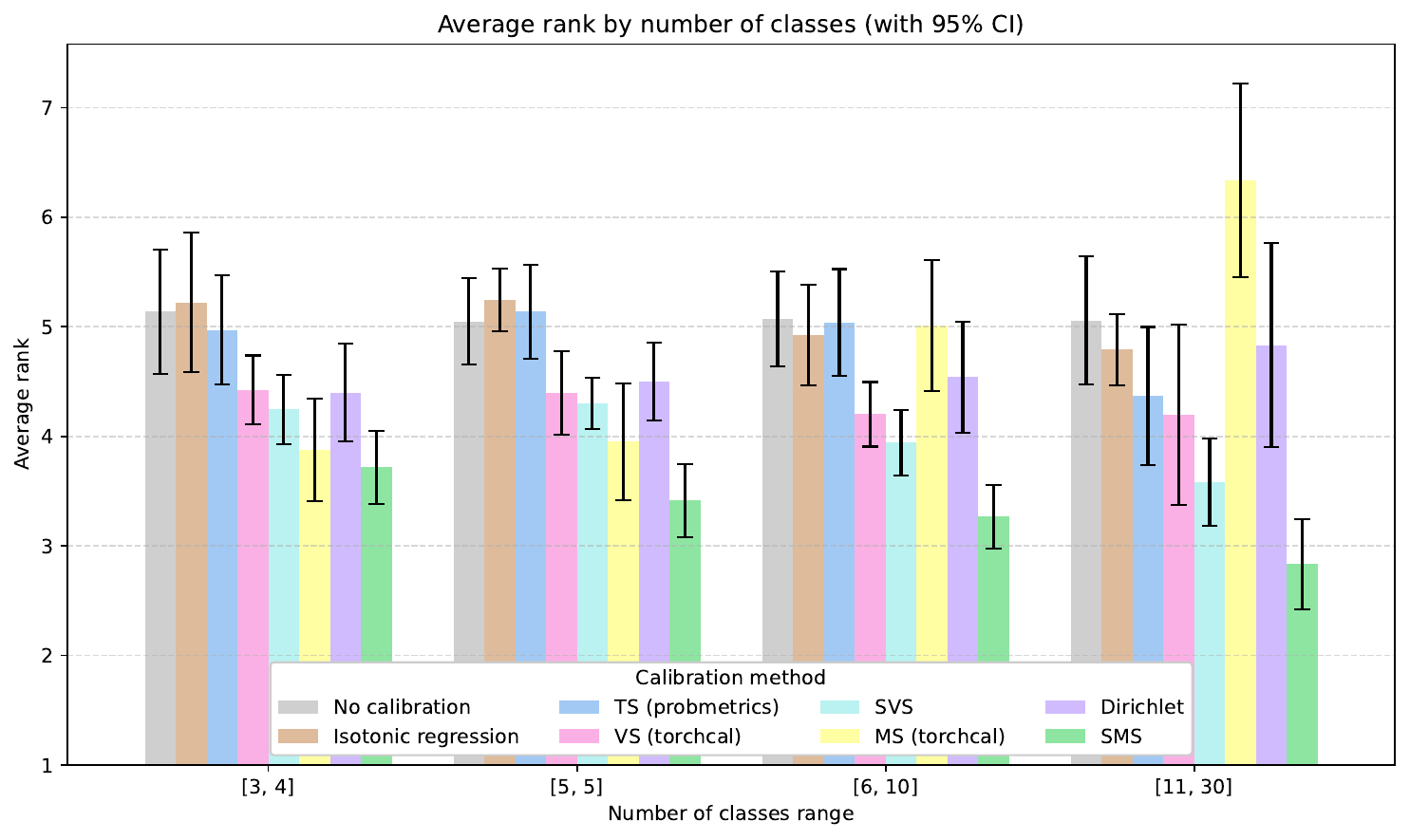}
    \caption{\emph{Average rank of multi-class calibration methods, per number of classes.}
    Datasets are grouped by growing number of classes, using equal mass binning with 4 bins.
    Each bin contains $65/4 \simeq 16$ datasets. 
    On each bin, we rank calibration methods by average test Brier score improvement on the dataset.
    We plot standard $95\%$ confidence intervals for the average ranks.}
\end{figure}

\begin{figure}[htbp]
    \centering
    \includegraphics[width=0.9\linewidth]{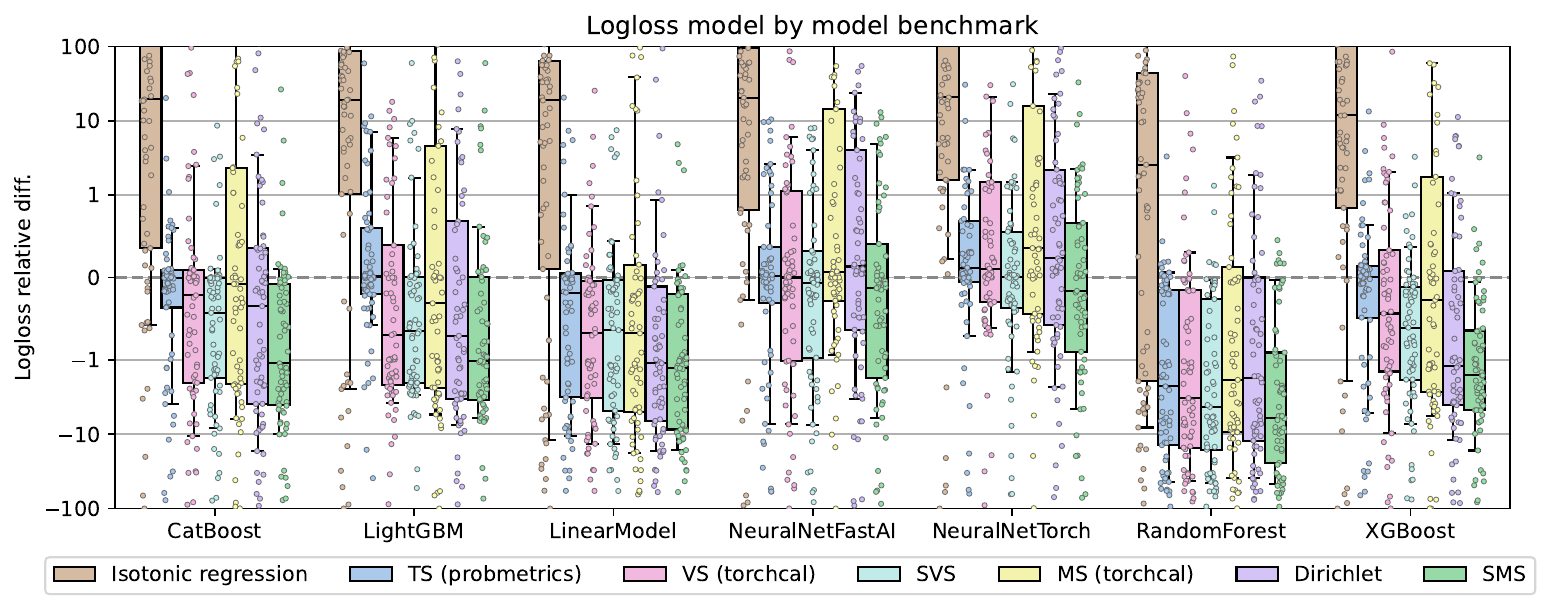}
    \caption{\emph{Relative differences in test logloss after recalibration (lower is better) for SVS and SMS, compared with other calibration methods, separately for every model in our benchmark.}
    Each dot represents the average relative loss difference obtained for one tabular dataset, over 21 experiments (7 models, 3 folds).
    Box-plots show the 10, 25, 50, 75, and 90\% quantiles. Relative differences (y-axis) are plotted using a log scale and clipped to -100\% loss (min) and +100\% loss (max).}
\end{figure}

\begin{figure}[htbp]
    \centering
    \includegraphics[width=0.9\linewidth]{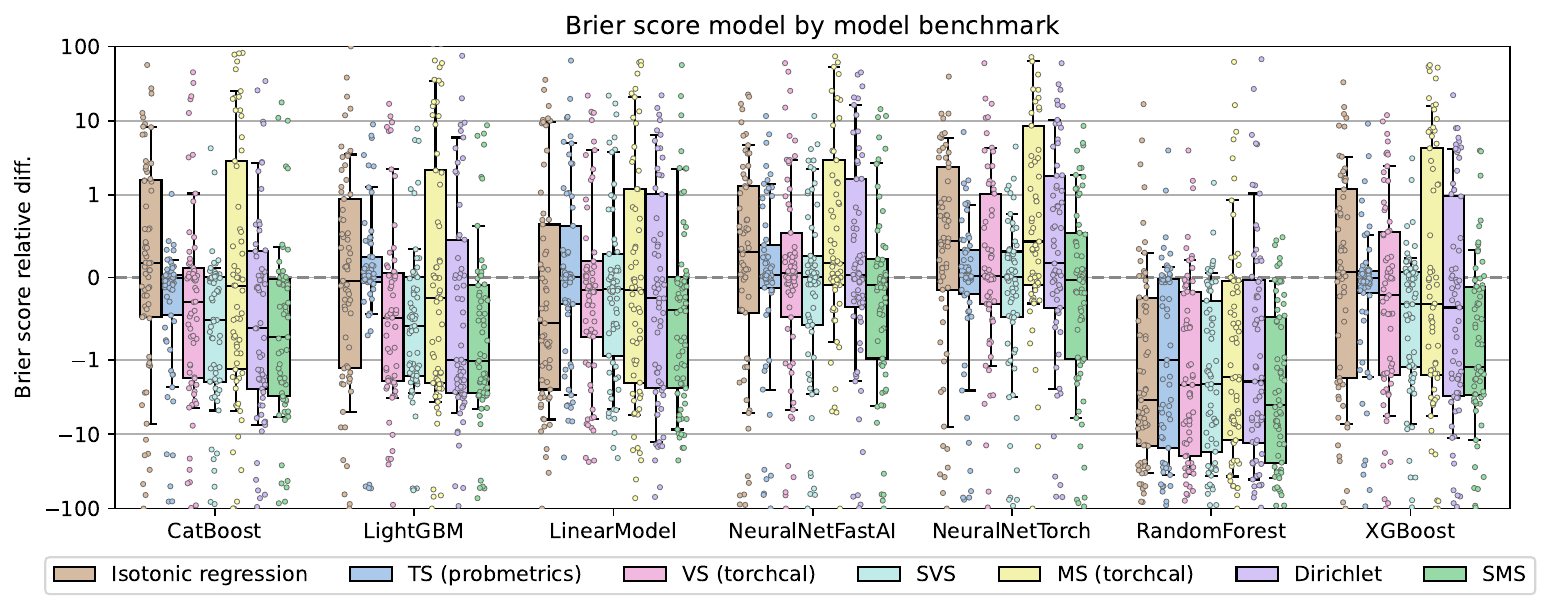}
    \caption{\emph{Relative differences in test Brier score after recalibration (lower is better) for SVS and SMS, compared with other calibration methods, separately for every model in our benchmark.}
    Each dot represents the average relative loss difference obtained for one tabular dataset, over 21 experiments (7 models, 3 folds).
    Box-plots show the 10, 25, 50, 75, and 90\% quantiles. Relative differences (y-axis) are plotted using a log scale and clipped to -100\% loss (min) and +100\% loss (max).}
\end{figure}

\begin{figure}[htbp]
    \centering
    \includegraphics[width=0.6\linewidth]{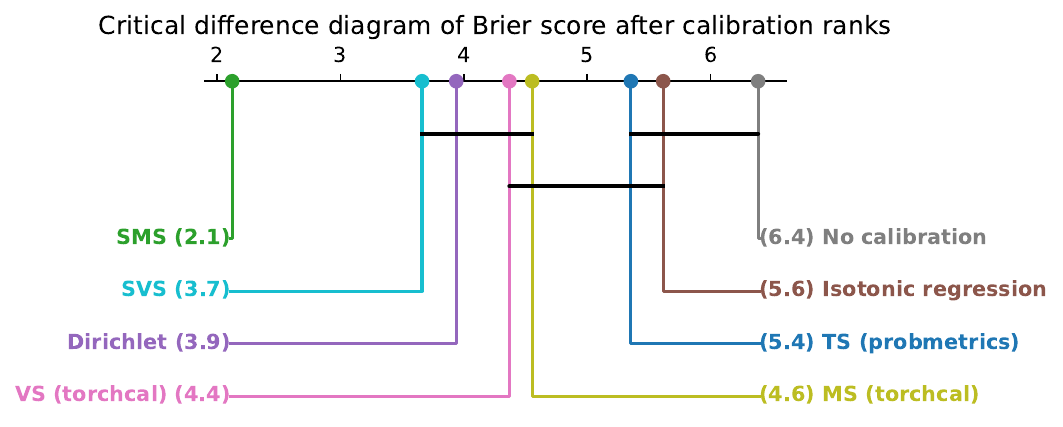}
    \caption{\emph{Critical differences diagram for Brier score after calibration.}
    Black lines indicate statistically indistinguishable methods. Number in parentheses indicate the average rank of each method on the 65 datasets (lower is better).}
\end{figure}

\end{document}